\documentclass{article}

\usepackage{microtype}
\usepackage{graphicx}
\usepackage{subfigure}
\usepackage{booktabs} 

\usepackage{hyperref}
\usepackage{stfloats}

\usepackage[accepted]{icml2025}

\usepackage{amsmath}
\usepackage{amssymb}
\usepackage{mathtools}
\usepackage{amsthm}

\usepackage{colortbl}
\definecolor{rank1}{HTML}{F7B79A}
\definecolor{deeper_rank1}{HTML}{FF9900}
\definecolor{rank2}{HTML}{87CEEB}
\definecolor{IFCD}{HTML}{D9ECFF}
\definecolor{Deeper_IFCD}{HTML}{7BABF8}
\usepackage{multirow}

\usepackage[capitalize,noabbrev]{cleveref}

\theoremstyle{plain}

\theoremstyle{definition}

\theoremstyle{remark}

\usepackage[textsize=tiny]{todonotes}

\icmltitlerunning{Mitigating Hallucinations in Large Vision-Language Models with Internal Fact-based Contrastive Decoding}

\begin{document}

\twocolumn[
\icmltitle{Mitigating Hallucinations in Large Vision-Language Models with Internal Fact-based Contrastive Decoding}



\icmlsetsymbol{Corresponding author}{\dag}

\begin{icmlauthorlist}
\icmlauthor{Chao Wang}{Corresponding author,future,ai}
\icmlauthor{Xuancheng Zhou}{future,ai}
\icmlauthor{Weiwei Fu}{future,ai}
\icmlauthor{Yang Zhou}{Corresponding author,ai,auto}
\end{icmlauthorlist}

\icmlaffiliation{future}{School of Future Technology, Shanghai University, Shanghai, 200444, China.}
\icmlaffiliation{ai}{Institute of Artificial Intelligence, Shanghai University, Shanghai, 200444, China.}
\icmlaffiliation{auto}{School of Mechatronic Engineering and Automation, Shanghai, 200444, China}

\icmlcorrespondingauthor{Chao Wang}{cwang@shu.edu.cn}
\icmlcorrespondingauthor{Yang Zhou}{saber\_mio@shu.edu.cn}

\icmlkeywords{Machine Learning, ICML}

\vskip 0.3in
]



\printAffiliationsAndNotice{\icmlCorrespondingAuthor} 

\begin{abstract}
Large Visual Language Models (LVLMs) integrate visual and linguistic modalities, exhibiting exceptional performance across various multimodal tasks. Nevertheless, LVLMs remain vulnerable to the issue of object hallucinations. Previous efforts to mitigate this issue focus on supervised fine-tuning (SFT) or incorporating external knowledge, both of which entail significant costs related to training and the acquisition of external data. To address these challenges, we propose a novel model-agnostic approach termed Internal Fact-based Contrastive Decoding (IFCD), designed to mitigate and suppress hallucinations during the inference process of LVLMs by exploiting the LVLMs' own hallucinations. IFCD is grounded in experimental observations that alterations to the LVLMs' internal representations tend to amplify hallucinations caused by language bias. By contrasting disturbed distribution, IFCD calibrates the LVLMs' output and effectively removes the hallucinatory logits from the final predictions. Experimental results validate that IFCD significantly alleviates both object-level and attribute-level hallucinations while achieving an average 9\% accuracy improvement on POPE and 8\% accuracy improvement on MME object hallucinations subset compared with direct decoding, respectively.
\end{abstract}

\section{Introduction}
In recent years, significant advancements have been made in developing large vision-language models (LVLMs) \cite{li2023blip, wen2024road}, which exhibit exceptional capabilities across a broad spectrum of tasks ~\cite{achiam2023gpt}. These models are increasingly viewed as a step toward achieving artificial general intelligence \cite{sanderson2023gpt}. LVLMs are capable of extracting intricate complex visual information and transforming it into continuous language representations for generation ~\cite{liu2024visual, zhu2024minigpt}. However, a critical challenge that persists with LVLMs is the phenomenon of hallucinations. Before the era of LVLMs, the natural language processing (NLP) community defined hallucinations as generated textual content that deviates from actuality \cite{ji2023survey, biten2022let}. With the advancements of LVLMs, a new form of hallucination has emerged, known as object hallucination. This refers to the generation that are inconsistent with visual input, and it becomes a significant issue that impedes the deployment of LVLMs in domains that require high reliability, particularly in risk-sensitive industries \cite{sahoo2024comprehensive}.

\begin{figure}
\begin{center}
\centerline{\includegraphics[width=1\linewidth]{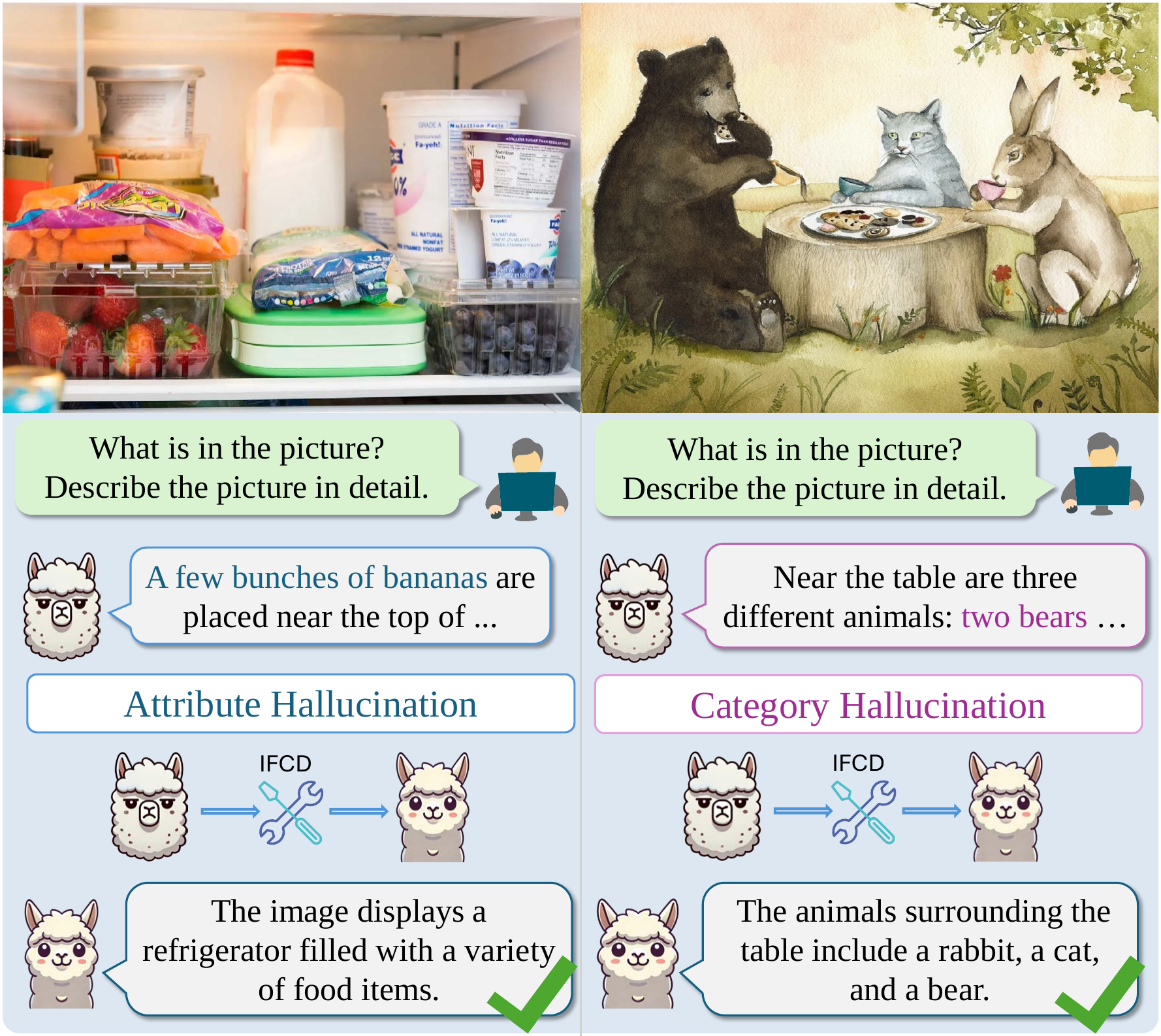}}
\caption{Cases of object hallucinations and effect of IFCD on LLaVA 1.5. Given two images, an LLaVA 1.5 outputs responses with attribute and category hallucinations which IFCD fixes.}
\label{fig: case of OH}
\vskip -0.4in
\end{center}
\end{figure}

Object hallucinations refer to the phenomenon where the language output generated by an LVLM fails to align with the visual input content ~\cite{song2024hscl, min2024mitigating}. In Figure \ref{fig: case of OH}, the LVLM incorrectly assumes that bananas are present in the refrigerator and even provides a false location of the bananas. At the same time, the LVLM accurately counts the number of animals but misclassifies an animal that is not a bear as a bear in another example. These two examples highlight the object hallucinations issue in LVLMs, which severely limits their applicability in domains where high accuracy is essential \cite{zhang2023siren, pal-etal-2023-med}. Therefore, addressing the object hallucinations is a pivotal step toward enhancing the reliability of LVLMs and expanding their potential applications.


To address the issue of object hallucinations in LVLMs, numerous works focus on incorporating external information to support fact-checking ~\cite{zhao2024mitigating, asai2023self}, thereby enhancing the factual accuracy of LVLMs through techniques such as self-evaluation \cite{singhal2024multilingual}. Additionally, improving LVLMs performance through preference fine-tuning is a prevalent strategy \cite{rafailov2024direct, stiennon2020learning}, aiming to align model outputs with human preferences and enhance model performance at a fine-grained level. While existing interventions for mitigating object hallucinations in LVLMs have shown effectiveness, the associated human and computational costs highlight the urgent need for simpler yet effective approaches.

To address these challenges, we propose Internal Fact-based Contrastive Decoding (IFCD), a novel model-agnostic approach that leverages hallucinations to mitigate further hallucination. IFCD can be seamlessly integrated into any open-source LVLM with minimal training required for the probe model. IFCD significantly enhances the truthfulness of LVLMs while reducing object hallucination. To assess the effectiveness of IFCD, we conduct experiments on two widely adopted LVLMs, LLaVA 1.5 \cite{liu2024visual} and InstructBLIP \cite{li-etal-2023-lavis}. Our evaluation using the Polling-based Object Probing Evaluation (POPE) \cite{li-etal-2023-evaluating} demonstrates that IFCD consistently outperforms baseline approaches, achieving up to a 9\% improvement in performance across all LVLMs. Additionally, IFCD enhances the overall perceptual capabilities of LVLMs, as evidenced by benchmarking on MME \cite{fu2023mme} and LLaVA-Bench \cite{liu2024visual}. In the text generation task, IFCD reduces the hallucinated object ratio by 5\%, while preserving the generated text's quality.

Concretely, our main \textbf{contributions} are as follows:
\textbf{1).}~We analyze the impact of editing internal representations on object hallucinations in LVLMs, with a particular focus on the effects of language bias.
\textbf{2).}~We introduce IFCD, a novel technique to calibrate LVLMs' output distribution and mitigate object hallucinations by contrasting the disturbed distribution derived from internal representation editing.
\textbf{3).}~IFCD demonstrates the effect in mitigating object hallucination, achieving 9\% and 8\% improvement on POPE and MME, respectively, and 13\% improvement in suppressing hallucinatory object effects while being more robust in the long text generation experiments.

\section{Related Work}
\subsection{Large Visual Language Models}
In recent years, large language models (LLMs) based on the Transformer architecture have achieved remarkable achievements in various fields, including Natural Language Processing (NLP), Machine Translation, and Computer Vision. ~\cite{zhao2023survey, achiam2023gpt, chiang2023vicuna}. Notably, with the introduction of multimodal models such as CLIP ~\cite{radford2021learning} and Vision Transformer ~\cite{dosovitskiy2021an}, LVLMs have been established through comprehensive pre-training processes that unify textual and visual modalities ~\cite{bai2023qwen, Ye_2024_CVPR}. Compared with traditional vision models, LVLMs adopted more advanced training paradigms ~\cite{wei2022finetuned, liu2024visual}. As a result, LVLMs demonstrated unique capabilities not present in traditional models \cite{yang2023mm}, including establishing application \cite{Ye_2024_CVPR}, and advanced mathematical reasoning \cite{pmlr-v202-driess23a}.

\subsection{Object Hallucination}
While LVLMs exhibited strong capabilities in addressing vision-language tasks, they were still significantly affected by object hallucinations ~\cite{li-etal-2023-evaluating}, generating content irrelevant to visual information. To identify the issue of object hallucinations in LVLMs, recent research has established specific indicators, such as Caption hallucinations Assessment with Image Relevance (CHAIR) ~\cite{rohrbach-etal-2018-object} and Sharpness \cite{chen2024context}. Additionally, advances have been made in locating the causes of hallucinations within LVLMs, including internal representation and attention patterns ~\cite{han2024semantic, mahaut-etal-2024-factual}. These metrics and locating approaches provided a multi-dimensional view to observe object hallucination. 

\begin{figure*}[ht!]
\vskip 0.2in
\begin{center}
\centerline{\includegraphics[width=\linewidth]{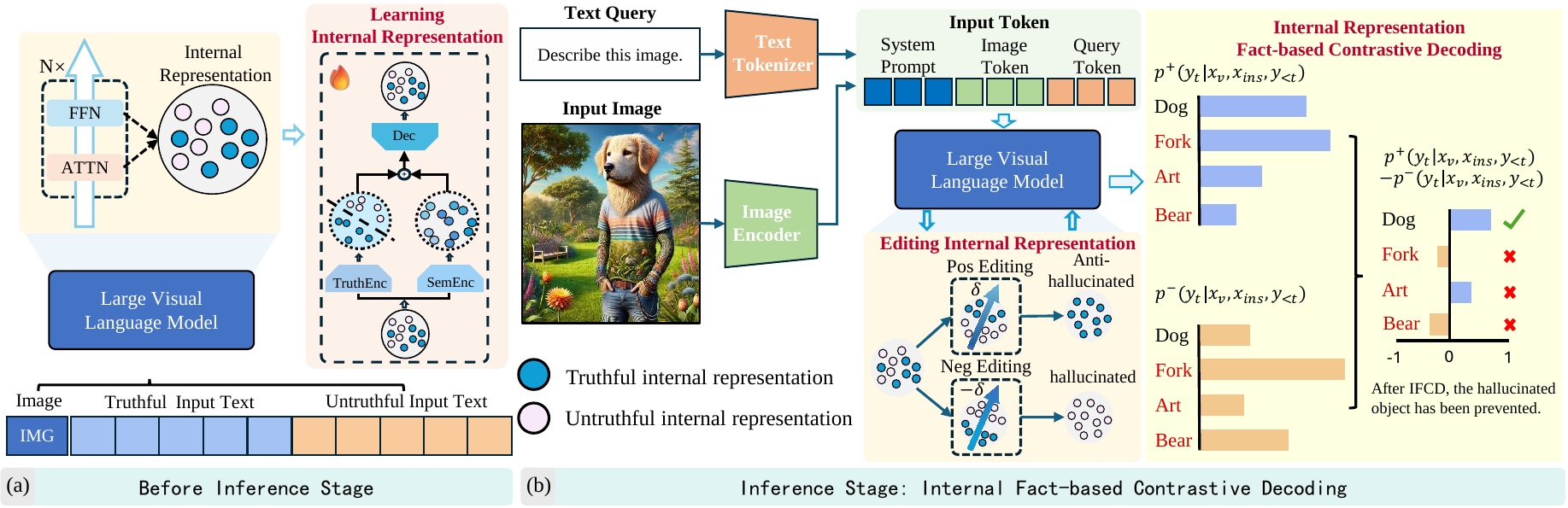}}
\caption{\textbf{An overview of IFCD.~} IFCD first edits the internal representation of the LVLMs to construct counterfactual logits for comparison by deliberately injecting hallucinations into the model trained by contrastive learning. These counterfactual logits are utilized to reveal potential hallucinatory tendencies of the LVLMs. Furthermore, the internal representation editing model is employed to actively attenuate a portion of the hallucinatory components within the LVLMs, thereby initiating an improvement in the factual accuracy of its outputs. This process effectively corrects the token from an erroneous token ``[Fork]'' to an accurate ``[Dog]''.}
\label{fig:pipe_line}
\vskip -0.3in
\end{center}
\end{figure*}

Addressing object hallucinations typically focused on direct suppression methods and fine-tuning, which involved actively limiting \cite{dhuliawala-etal-2024-chain}, correcting hallucinated outputs \cite{hu-etal-2024-knowledge} and RLHF to refine LVLMs ~\cite{ye2024mplug}. These approaches often constructed enhanced datasets for fine-tuning or training LVLMs. As the parameter scales of LVLMs continued to increase, these challenges became even more pronounced~\cite{Ye_2024_CVPR, zhu2024minigpt}. To tackle these issues, IFCD actively induces hallucinations into the model's output and leverages them as counterexamples to refine the model's final responses, thereby reducing the likelihood of hallucinations in final outputs. IFCD leverages hallucinated outputs as improving opportunities, offering a novel way to mitigate object hallucinations without high computation costs.

\section{Method}
\subsection{Overview}
In this section, we propose IFCD to mitigate object hallucinations in LVLMs effectively. IFCD constructs two distributions with a truthfulness gap via internal representation editing and mitigates object hallucinations using contrastive decoding to subtract hallucinatory distributions. Section \ref{3.2} details the editing process, while Section \ref{3.3} explains the contrastive decoding mechanism. The overall framework is illustrated in Figure \ref{fig:pipe_line}.

\subsection{Amplifying Object hallucinations via Internal Representation Editing} \label{3.2}
\textbf{Induction of Object hallucinations   } A substantial proportion of object hallucinations arises from statistical bias \cite{zhao2024mitigating}, for which contrastive decoding has emerged as an effective countermeasure \cite{chenhalc, min2024mitigating}. The methodology originates from contrastive decoding \cite{li-etal-2023-contrastive}, which subtracts logits that deviate from expected distributions, thereby enhancing the performance of LVLMs. Existing approaches reveal LVLMs' tendency toward object hallucinations by confusing the visual input \cite{leng2024mitigating} or instructing LVLMs to make incorrect decisions \cite{wang-etal-2024-mitigating} and mitigate hallucinations via contrastive decoding. Therefore, a critical question arises: \textit{How can we create hallucination-inducing samples that reflect token distribution errors while generating significant hallucination?}

We argue that previous methods relying on distracting information are suboptimal for inducing object hallucinations in LVLMs. These approaches primarily highlight the models' reactions to specific commands or perturbations. Although these responses approximate object hallucination, they stem from exogenous factors rather than inherent model errors and hardly fully represent LVLMs' object hallucination. To better simulate logits indicative of object hallucination, we propose intervening in the internal representation of LVLMs during inference.

\textbf{Introduction of Internal Representation Editing } Intuitively, the internal representations from the attention and feedforward layers directly contribute to the inference process, thereby influencing the model's output. This section delves into an analysis aiming to validate the hypotheses that editing internal representation can amplify object hallucinations in LVLMs. There are various methods to intervene in internal representations to alter output truthfulness \cite{pan2024towards, chen2024incontext}. Considering compatibility and availability, we propose using TruthX \cite{truthx} to edit the internal representations of LVLMs. TruthX is an autoencoder-based model comprising two encoders, $\mathrm{TruthEnc(\cdot)}$ and $\mathrm{SemEnc(\cdot)}$, and a decoder $\mathrm{Dec(\cdot)}$, all implemented with multi-layer perceptions (MLPs). Two encoders map LVLMs' internal representations $x$ as follows:
\begin{equation}
    h_{\textit{truth}}=\mathrm{TruthEnc}(x), h_{\textit{sem}}=\mathrm{SemEnc}(x),
\end{equation}
where $h_{\textit{truth}}$, $h_{\textit{sem}}\in \mathbb{R}^{d_{latent}}$ are the latent representations in the latent spaces of $\mathrm{TruthEnc(\cdot)}$ and $\mathrm{SemEnc(\cdot)}$, $d_{latent}$ is the dimension of latent space. Then decoder $\mathrm{Dec(\cdot)}$ reconstructs LVLM internal representation:
\begin{equation}
x^\prime=\mathrm{Dec}(h_{\textit{sem}}+\mathrm{Attn}(h_\textit{sem},h_\textit{truth})),
\end{equation}
where $x^\prime$ is the reconstructed representation and $\mathrm{Attn(\cdot)}$ is an attention operation from semantic latent representation to truthful latent representation.

\textbf{Internal Representation Editing Amplifies Object hallucinations  }
Through contrastive learning, the truthfulness of the LVLMs' internal representation can be discerned within $\mathrm{TruthEnc(\cdot)}$. The editing procedure is subsequently determined by the relative positions of the central point of the latent space vectors corresponding to truthful and untruthful representations mapped by $\mathrm{TruthEnc(\cdot)}$. Formally, the direction $\delta \in \mathbb{R}^{d_{latent}}$ of internal representation editing can be defined as follows:
\begin{equation}
\delta = \overline{\mathcal{H}}^{pos}_{truth} - \overline{\mathcal{H}}^{neg}_{truth},
\end{equation}
where $\overline{\mathcal{H}}^{pos}_{truth}$ and $\overline{\mathcal{H}}^{neg}_{truth}$ denotes the average position of mappings of truthful and untruthful representations within the latent space of $\mathrm{TruthEnc(\cdot)}$. Note that the determination of editing direction $\delta$ happens during the training stage of TruthX. Adjustments to truthfulness can be reversed if the opposite direction $-\delta$ is used.

In the inference process of the LVLMs, TruthX maps and edits the internal representation in the latent space of $\mathrm{TruthEnc(\cdot)}$, then reconstructs the internal representation. Through contrasting the difference between the original and reconstructed internal representation, the content of the modifications to the internal representation $\Delta \in \mathbb{R}^{d_{model}}$ can be effectively derived, where $d_{model}$ refers to the dimension of LVLMs internal representations.

Then, the internal representation $x$ of LVLMs is edited as follows formally:
\begin{equation}
\hat{x} = x + \gamma \times \Delta,
\end{equation}
where $\gamma$ denotes the editing strength and $\hat{x}$ is the reconstructed internal representation of LVLMs. In practice, it is not necessary to edit all attention and feedforward layers. Modifying the layers that are sensitive to factual differences alone can yield substantial change in object hallucination.

\begin{figure}[h!]
\vskip 0.2in
\begin{center}
 \centerline{\includegraphics[width=1\linewidth]{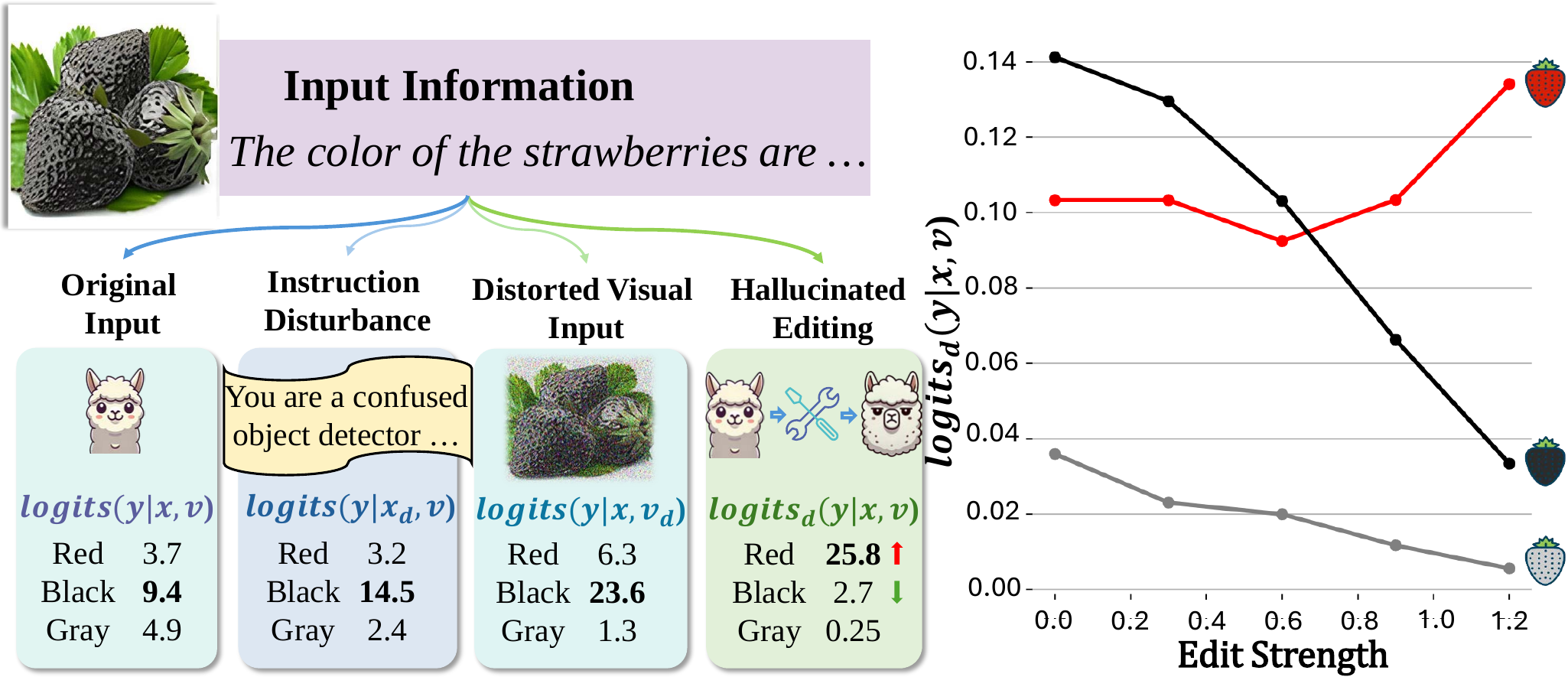}}
\caption{An illustration of editing internal representation amplifying language priors. Given an image depicting three black strawberries, LVLMs assign more preference for more conventional strawberry color, such as ``red'', with increasing editing strength.}
\label{fig:hallu-induced}
\end{center}
\vskip -0.3in
\end{figure}

We compare the effect of editing the internal representation of LVLMs to expose object hallucinations preference with two alternative interference methods, using the case of recognizing black strawberries on LLaVA 1.5 in Figure \ref{fig:hallu-induced}. For Instruction Dirturbance, we follow Instruction Contrastive Decoding \cite{wang-etal-2024-mitigating}, utilizing the prompt ``\texttt{You are a confused object detector.}'' to induce hallucination. For visual disturbance, we adopt the methodology from Visual Contrastive Decoding \cite{leng2024mitigating}, setting the noise step parameter to 400, which controls the scale of noise added. Figure \ref{fig:hallu-induced} presents editing the internal representation enables the LVLMs to disregard visual information and disproportionately rely on language priors in its decision-making process. Additionally, increasing the strength of the internal representation modification can further expose the statistical bias in the responses.
\subsection{Internal Fact-based Contrastive Decoding} \label{3.3}
\textbf{Contrasting the Predictions with Disturbance    } The findings from our previous analyses substantiate the hypothesis that \textit{manipulating the internal representations of LVLMs can exacerbate object hallucination, thereby making hallucinatory content more strongly reflective of untruthful information}. A promising approach to mitigate object hallucinations of LVLMs is to directly subtract the logits associated with hallucinatory content from the final output logits, thereby enabling a more targeted mitigation of object hallucination. Building upon this view, we introduce IFCD aimed at alleviating hallucinations during LVLMs inference.

Drawing from the concept of contrastive decoding \cite{li-etal-2023-contrastive}, which enhances the overall quality of LLM outputs by comparing logits from two models with performance discrepancies, we contrast the generations from the hallucination-inducing and hallucination-suppressing models to improve final performance. Specifically, as demonstrated in Figure \ref{fig:pipe_line}, given the visual features $X_v$ extracted from the visual encoder and the textual query $X_q$, our method computes two distinct token distributions: the first distribution $P^+$ is derived from the LVLM edited for anti-hallucination. In contrast, the second distribution $P^-$ is generated from the LVLM after modification of its internal representations for hallucination-inducing. In contrast to the regular approach of selecting the token with the highest probability, our method relies on the two token distributions, $P^+$ and $P^-$, to inform the final token selection. The contrasting token probability distribution is computed by evaluating the difference between $P^+$ and $P^-$ as follows: 
\begin{equation}
\begin{split}
p_{\text{IFCD}}(y_t|x_v, x_q) =  \sigma\left((1 + \alpha)p^{+}(y_t|\ast) - \alpha p^{-}(y_t|\ast)\right),
\end{split}
\label{eq:IFCD}
\end{equation}
where $\ast$ denotes visual and textual information given to LVLMs as well as previously generated tokens, while $\alpha$ is employed to control the contrast strength, and $y_t$ refers to the token generated in $t$-th position.

\textbf{Adaptive Plausibility Constraints   } The fundamental principle of IFCD is to prioritize the selection of tokens with high probabilities as predicted by the LVLMs, while simultaneously imposing penalties on those tokens associated with hallucinatory logits. However, this approach risks inadvertently affecting tokens that are correctly identified both under standard conditions and hallucinated content. Imposing penalties on these tokens may inadvertently reward unreliable tokens that should not be prioritized, potentially distorting the LVLMs’ output. To mitigate this issue, we introduce constraints on the scope of influence exerted by IFCD, drawing upon adaptive plausibility constraints that are employed in the open-ended text generation realm ~\cite{li-etal-2023-contrastive}.
\begin{equation}
\begin{split}
y_t\sim P_{\text{IFCD}} \textit{, s.t. } y_t \in \mathcal{V}_{head}(y_{<t}),
\end{split}
\end{equation}
\begin{equation}
\begin{split}
\mathcal{V}_{head}(y_{<t})=  \{y_t \in \mathcal{V}: p(y_t|\ast) \geq \beta \max_{[\text{T}]}p([\text{T}]|\ast) \},
\end{split}
\end{equation}
where $[\text{T}]$ refers to the candidate tokens, while the pivotal hyperparameter $\beta$ serves to regulate the truncation strength of the logits, thereby determining the tokens affected during the contrastive decoding process. This parameter is crucial for constraining the impact on irrelevant tokens, thereby ensuring the robustness of the contrastive decoding process.

\section{Experiments} \label{4}

\begin{table*}[ht!]
\caption{Results on object hallucinations benchmark POPE. Regular denotes direct sampling, whereas ICD and VCD are two baselines for comparison, and IFCD is our proposed decoding. The best performance is marked by \textbf{bold}, and performance of IFCD is marked by \textcolor{Deeper_IFCD}{cyan}.}
\begin{center}
\begin{small}
\begin{sc}

\resizebox{\textwidth}{!}{
\begin{tabular}{lllcccc|cccc}
\hline
                          &                               &                            & \multicolumn{4}{c|}{LLaVA 1.5}                                                                                                                           & \multicolumn{4}{c}{InstructBLIP}                                                                                                                         \\ \cline{4-11} 
\multirow{-2}{*}{Dataset} & \multirow{-2}{*}{Setting}     & \multirow{-2}{*}{Decoding} & \multicolumn{1}{l}{Accuracy}           & \multicolumn{1}{l}{Precision}          & \multicolumn{1}{l}{Recall}    & \multicolumn{1}{l|}{F1 Score}          & \multicolumn{1}{l}{Accuracy}           & \multicolumn{1}{l}{Precision}          & \multicolumn{1}{l}{Recall}    & \multicolumn{1}{l}{F1 Score}           \\ \hline
                          &                               & Regular                    & 83.29                                  & 92.13                                  & 72.80                         & 81.33                                  & 80.71                                  & 81.67                                  & 79.19                         & 80.41                                  \\
                          &                               & ICD                        & 84.23                                  & \textbf{95.08}                         & 72.20                         & 82.08                                  & 83.50                                  & 87.69                                  & 77.93                         & 82.52                                  \\
                          & \multirow{-2}{*}{Random}      & VCD                        & 87.73                                  & 91.42                                  & \textbf{83.28}                & 87.16                                  & 84.53                                  & 88.55                                  & \textbf{79.32}                & 83.68                                  \\
                          &                               & \textbf{IFCD (Ours)}       & \cellcolor{IFCD}\textbf{89.17} & \cellcolor{IFCD}94.54          & \cellcolor{IFCD}83.13 & \cellcolor{IFCD}\textbf{88.47} & \cellcolor{IFCD}\textbf{85.56} & \cellcolor{IFCD}\textbf{97.09} & \cellcolor{IFCD}73.33 & \cellcolor{IFCD}\textbf{83.75} \\ \cline{3-11} 
                          &                               & Regular                    & 81.88                                  & 88.93                                  & 72.8                          & 80.06                                  & 78.22                                  & 77.87                                  & 78.85                         & 78.36                                  \\
                          &                               & ICD                        & 82.73                                  & 91.47                                  & 72.20                         & 80.70                                  & 79.40                                  & 80.28                                  & 77.93                         & 79.09                                  \\
\multirow{-2}{*}{MSCOCO}  & \multirow{-2}{*}{Popular}     & VCD                        & 85.38                                  & 86.92                                  & \textbf{83.28}                & 85.06                                  & 81.47                                  & 82.89                                  & \textbf{79.32}                & 81.07                                  \\
                          &                               & \textbf{IFCD (Ours)}       & \cellcolor{IFCD}\textbf{88.10} & \cellcolor{IFCD}\textbf{93.13} & \cellcolor{IFCD}82.27 & \cellcolor{IFCD}\textbf{87.36} & \cellcolor{IFCD}\textbf{83.27} & \cellcolor{IFCD}\textbf{91.93} & \cellcolor{IFCD}72.93 & \cellcolor{IFCD}\textbf{81.34} \\ \cline{3-11} 
                          &                               & Regular                    & 78.96                                  & 83.06                                  & 72.75                         & 77.57                                  & 75.84                                  & 74.30                                  & 79.03                         & 76.59                                  \\
                          &                               & ICD                        & 80.23                                  & 85.96                                  & 72.27                         & 78.52                                  & 77.70                                  & 77.57                                  & 77.93                         & 77.75                                  \\
                          & \multirow{-2}{*}{Adversarial} & VCD                        & 80.88                                  & 79.45                                  & \textbf{83.29}                & 81.33                                  & 79.56                                  & 79.67                                  & \textbf{79.39}                & 79.52                                  \\
                          &                               & \textbf{IFCD (Ours)}       & \cellcolor{IFCD}\textbf{85.17} & \cellcolor{IFCD}\textbf{86.76} & \cellcolor{IFCD}83.00 & \cellcolor{IFCD}\textbf{84.84} & \cellcolor{IFCD}\textbf{82.23} & \cellcolor{IFCD}\textbf{89.47} & \cellcolor{IFCD}73.07 & \cellcolor{IFCD}\textbf{80.44} \\ \hline
                          &                               & Regular                    & 83.45                                  & 87.24                                  & 78.36                         & 82.56                                  & 80.91                                  & 77.97                                  & 86.16                         & 81.86                                  \\
                          &                               & ICD                        & 86.13                                  & \textbf{91.44}                         & 79.73                         & 85.19                                  & 82.83                                  & 82.42                                  & 83.46                         & 82.94                                  \\
                          & \multirow{-2}{*}{Random}      & VCD                        & 86.15                                  & 85.18                                  & \textbf{87.53}                & 86.34                                  & 84.11                                  & 82.21                                  & \textbf{87.05}                & 84.56                                  \\
                          &                               & \textbf{IFCD (Ours)}       & \cellcolor{IFCD}\textbf{87.30} & \cellcolor{IFCD}89.54          & \cellcolor{IFCD}84.47 & \cellcolor{IFCD}\textbf{86.93} & \cellcolor{IFCD}\textbf{85.83} & \cellcolor{IFCD}\textbf{93.10} & \cellcolor{IFCD}77.40 & \cellcolor{IFCD}\textbf{84.58} \\ \cline{3-11} 
                          &                               & Regular                    & 79.90                                  & 80.85                                  & 78.36                         & 79.59                                  & 76.19                                  & 72.16                                  & 85.28                         & 78.17                                  \\
                          &                               & ICD                        & 82.5                                   & \textbf{84.40}                         & 79.73                         & 82.00                                  & 77.23                                  & 74.21                                  & 83.46                         & 78.56                                  \\
\multirow{-2}{*}{A-OKVQA} & \multirow{-2}{*}{Popular}     & VCD                        & 81.85                                  & 78.60                                  & \textbf{87.53}                & 82.82                                  & 79.80                                  & 76.00                                  & \textbf{87.05}                & 81.15                                  \\
                          &                               & \textbf{IFCD (Ours)}       & \cellcolor{IFCD}\textbf{84.10} & \cellcolor{IFCD}84.17          & \cellcolor{IFCD}84.00 & \cellcolor{IFCD}\textbf{84.08} & \cellcolor{IFCD}\textbf{83.17} & \cellcolor{IFCD}\textbf{87.21} & \cellcolor{IFCD}77.73 & \cellcolor{IFCD}\textbf{82.20} \\ \cline{3-11} 
                          &                               & Regular                    & 74.04                                  & 72.08                                  & 78.49                         & 75.15                                  & 70.71                                  & 65.91                                  & 85.83                         & 75.56                                  \\
                          &                               & ICD                        & 76.70                                  & \textbf{75.08}                         & 79.93                         & 77.43                                  & 72.20                                  & 68.07                                  & 83.6                          & 75.04                                  \\
                          & \multirow{-2}{*}{Adversarial} & VCD                        & 74.97                                  & 70.01                                  & \textbf{87.36}                & 77.73                                  & 74.33                                  & 69.46                                  & \textbf{86.87}                & 77.19                                  \\
                          &                               & \textbf{IFCD (Ours)}       & \cellcolor{IFCD}\textbf{77.67} & \cellcolor{IFCD}74.73          & \cellcolor{IFCD}83.60 & \cellcolor{IFCD}\textbf{78.91} & \cellcolor{IFCD}\textbf{77.97} & \cellcolor{IFCD}\textbf{77.99} & \cellcolor{IFCD}77.93 & \cellcolor{IFCD}\textbf{77.96} \\ \hline
                          &                               & Regular                    & 83.73                                  & 87.16                                  & 79.12                         & 82.95                                  & 79.65                                  & 77.14                                  & 84.29                         & 80.56                                  \\
                          &                               & ICD                        & 86.10                                  & 90.38                                  & 80.80                         & 85.32                                  & 82.30                                  & 81.94                                  & 82.87                         & 82.40                                  \\
                          & \multirow{-2}{*}{Random}      & VCD                        & 86.65                                  & 84.85                                  & \textbf{89.24}                & 86.99                                  & 83.69                                  & 81.84                                  & \textbf{86.61}                & \textbf{84.16}                         \\
                          &                               & \textbf{IFCD (Ours)}       & \cellcolor{IFCD}\textbf{87.97} & \cellcolor{IFCD}\textbf{90.94} & \cellcolor{IFCD}84.33 & \cellcolor{IFCD}\textbf{87.51} & \cellcolor{IFCD}\textbf{84.77} & \cellcolor{IFCD}\textbf{92.50} & \cellcolor{IFCD}75.67 & \cellcolor{IFCD}83.24          \\ \cline{3-11} 
                          &                               & Regular                    & 78.17                                  & 77.64                                  & 79.12                         & 78.37                                  & 73.87                                  & 69.63                                  & 84.69                         & 76.42                                  \\
                          &                               & ICD                        & 80.00                                  & \textbf{79.53}                                  & 80.80                         & 80.16                                  & 74.70                                  & 71.23                                  & 82.87                         & 76.61                                  \\
\multirow{-2}{*}{GQA}     & \multirow{-2}{*}{Popular}     & VCD                        & \textbf{80.73}                         & 76.26                                  & \textbf{89.24}                & \textbf{82.24}                         & 78.57                                  & 74.62                                  & \textbf{86.61}                & \textbf{80.17}                         \\
                          &                               & \textbf{IFCD (Ours)}       & \cellcolor{IFCD}79.76          & \cellcolor{IFCD}77.61 & \cellcolor{IFCD}83.67 & \cellcolor{IFCD}80.52          & \cellcolor{IFCD}\textbf{80.13} & \cellcolor{IFCD}\textbf{82.90} & \cellcolor{IFCD}75.93 & \cellcolor{IFCD}79.26          \\ \cline{3-11} 
                          &                               & Regular                    & 75.08                                  & 73.19                                  & 79.16                         & 76.06                                  & 70.56                                  & 66.12                                  & 84.33                         & 74.12                                  \\
                          &                               & ICD                        & 77.47                                  & 76.08                                  & 80.13                         & 78.05                                  & 72.27                                  & 68.43                                  & 82.67                         & 74.88                                  \\
                          & \multirow{-2}{*}{Adversarial} & VCD                        & 76.09                                  & 70.83                                  & \textbf{88.75}                & 78.78                                  & 75.08                                  & 70.59                                  & \textbf{85.99}                & 77.53                                  \\
                          &                               & \textbf{IFCD (Ours)}       & \cellcolor{IFCD}\textbf{79.03} & \cellcolor{IFCD}\textbf{76.57} & \cellcolor{IFCD}83.67 & \cellcolor{IFCD}\textbf{79.96} & \cellcolor{IFCD}\textbf{78.00} & \cellcolor{IFCD}\textbf{79.49} & \cellcolor{IFCD}75.47 & \cellcolor{IFCD}\textbf{77.62} \\ \hline
\end{tabular}
}
\end{sc}
\end{small}
\end{center}
\vskip -0.1in
    \label{tab:POPE_EXP}
\end{table*}
\subsection{Experiments Settings}
\textbf{Benchmarks  } 
\textbf{POPE} \cite{li-etal-2023-evaluating}  comprises 1,500 images from three sources—MSCOCO, A-OKVQA, and GQA—along with 27,000 associated questions, focusing on detecting object existence hallucination. It also incorporates three sampling methods—random, popular, and adversarial—to evaluate the robustness of LVLMs against object hallucinations driven by statistical biases.
\textbf{MME}~\cite{fu2023mme} provides a total of 14 perception and cognition tasks. Since the cognitive task is related to the language decoder's reasoning capacity rather than visual content comprehension, we choose the perceptual subset of MME as the benchmark.  
Among these tasks, \textit{existence, count, position, and color} tasks are specifically designed as hallucinations discrimination benchmarks. \textbf{MSCOCO} \cite{MSCOCO} is a widely used computer vision benchmark, which contains more than 200,000 manually labeled high-quality and complex image-captions pairs, therefore very suitable for evaluating the object hallucinations problem. We randomly selected 500 images from MSCOCO to validate the long text generation ability of our method. \textbf{LLaVA-Bench} \cite{liu2024visual} contains 24 images with 60 questions, and the images cover a range of content such as portraits, landscapes, and enigmatic causes. We conduct a case study with this dataset to qualitatively demonstrate the effectiveness of our proposed IFCD.

\textbf{Metrics } The POPE evaluation pivots four key metrics: Accuracy, Precision, Recall, and the F1 score. To MME, we quantify performance via official implementation, the combined metric of accuracy and accuracy+. MSCOCO tests text generation capacity, which is quantified by BLEU \cite{papineni-etal-2002-bleu} and CHAIR \cite{rohrbach-etal-2018-object}. Specifically, CHAIR contains two sub-metrics $\text{CHAIR}_i$ and $\text{CHAIR}_s$, for object-focused and sentence-focused levels respectively. Formally, $\text{CHAIR}_i$ and $\text{CHAIR}_s$ could be described as follows:
\begin{equation}
\begin{split}
    &\text{CHAIR}_i = \frac{|\{\text{hallucinated objects}\}|}{|\{\text{all objects mentioned}\}|}, \\
    \text{CHAIR}_s& = \frac{|\{\text{sentences with hallucinated object}\}|}{|\{\text{all sentences}\}|},
\end{split}
\end{equation}
\textbf{LVLMs Baselines } We evaluate the effectiveness of IFCD on two popular LVLMs, LLaVA 1.5 \cite{liu2024visual} and InstructBLIP \cite{li-etal-2023-lavis}, configured with Vicuna 7B as the language decoder. Additionally, we reproduce the widely recognized Visual Contrastive Decoding (VCD) \cite{leng2024mitigating} and Instruction Contrastive Decoding (ICD) \cite{wang-etal-2024-mitigating} as comparisons with IFCD. Through comprehensive experiments, we demonstrate that IFCD is model-agnostic and can be seamlessly integrated with various LVLM architectures.

\textbf{Implementation Details  } \label{implementation_detail} In the experiments, we follow all hyperparameter settings in Appendix \ref{apex: param} unless otherwise noted. We use 300 MSCOCO images paired with both correct and incorrect responses as the training dataset for TruthX. We demonstrate and discuss the performance of IFCD with different training sizes in Appendix \ref{training_size_apex}. When adjusting the internal representations of LVLMs, we modify only the top 15 most important layers, with the editing strength $s=0.5$. It is worth noting that all experiments were conducted on a single RTX 3090 (24GB), highlighting the low cost of implementing IFCD.


\subsection{Experimental Results}

\begin{figure}[h]
\begin{center}
\centerline{\includegraphics[width=0.78\linewidth]{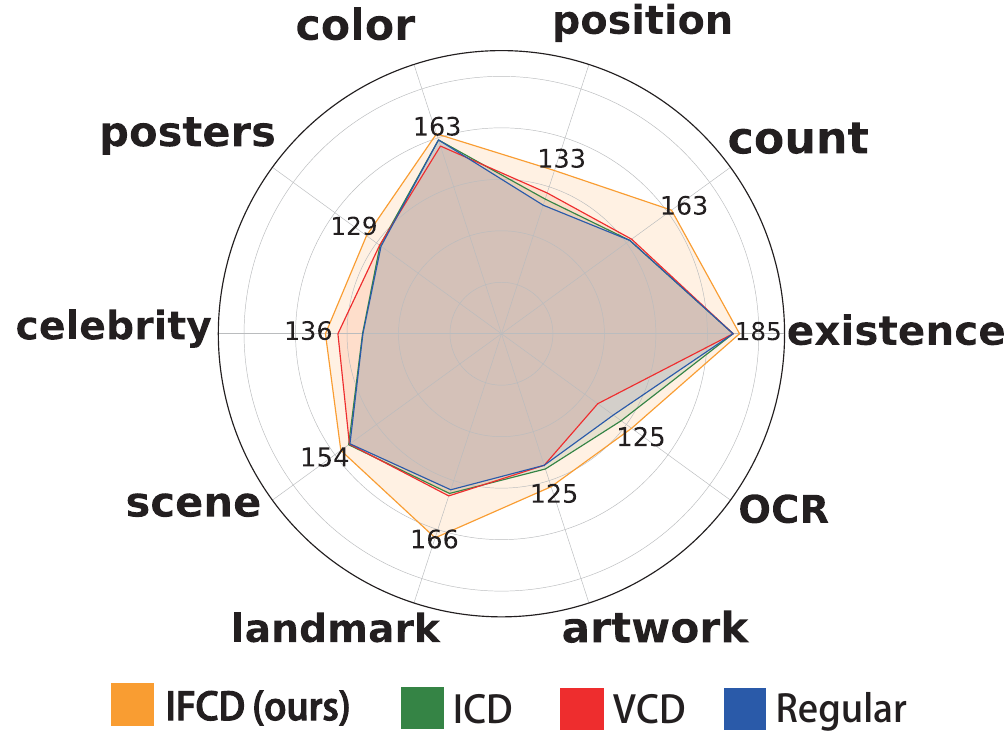}}
\caption{Perception subset of MME results on LLaVA 1.5. Regular denotes the direct sampling method, whereas ICD refers to the Instruction Contrastive Decoding, VCD refers to the Visual Contrastive Decoding baseline and IFCD is a sampling from our proposed contrastive decoding.}
\label{fig:mme_tasks}
\end{center}
\vskip -0.3in
\end{figure}

\textbf{Results on POPE~}
We use POPE to assess whether LVLMs misinterpret image objects. By applying different sampling methods in POPE, we can further evaluate the effectiveness of various methods in addressing statistical bias. As shown in Table \ref{tab:POPE_EXP}, two LVLMs employed IFCD achieve average accuracy improvements of up to 6.22 and 7.69, respectively, and F1 score improvements of up to 7.14 and 7.62, respectively, compared to the direct decoding, demonstrating the effectiveness of IFCD. A notable observation is that \textit{the different POPE setting varies the performance of all LVLMs with all methods, which could confirm the research that statistic bias is a cause of object hallucination} \cite{zhou2023analyzing}. However, among the four decoding methods, the performance degradation from the random setting to the adversarial setting is smaller for IFCD compared to the others. While direct decoding degrades by 7.8, and VCD and ICD degrade by 8.7 and 8.1, respectively, IFCD performs only a 6.8 degradation on average. The performance improvement and smaller performance degradation scale show the effectiveness and stability of IFCD. 

\textbf{Results on MME~} 
To evaluate LVLMs on diverse perceptual tasks, complement POPE’s focus on object existence, and enable a broader performance assessment, we experiment on MME. As shown in Figure \ref{fig:mme_tasks}, there is a general improvement in perception-related tasks with the application of IFCD. Table \ref{tab:subset_mme_blip} provides a more detailed comparison within the subset of object hallucination tasks in the MME. IFCD proves effective in attenuating the overall hallucination rate of the LVLMs, resulting in a 7.6\% and 13.8\% increase in the total score of InstructBLIP and LLaVA 1.5 respectively. Specifically, there is a notable increase in count and color tasks, suggesting that LVLMs are particularly susceptible to mistakes in these areas. Editing the internal representations proves to be an effective method for inducing hallucinations in these contexts. Therefore these errors can be mitigated through contrastive decoding to address object hallucination. In contrast, the position score is relatively low across four metrics, with minimal uplift from IFCD, suggesting the relatively weak ability of LVLMs in position reasoning. We demonstrate and discuss the performance of MME in more detail in Appendix \ref{appendix_mme}.
\begin{table}[h]
\vskip -0.1in
\caption{Results on object hallucination subset of MME on three decoding methods. The champion is marked by \textbf{bold} and \textbf{\textcolor{rank1}{orange}}, and the runner-up is marked by \textcolor{rank2}{cyan}.}
\begin{center}
\begin{small}
\begin{sc}
\resizebox{0.48\textwidth}{!}{
\begin{tabular}{llcccc|c}
\hline
Model                          & Decoding & \multicolumn{2}{c}{Object-level}                                              & \multicolumn{2}{c|}{Attribute-level}                                            & \multicolumn{1}{l}{Total Scores}       \\
                               &          & \multicolumn{1}{l}{Existence}        & \multicolumn{1}{l}{Count}              & \multicolumn{1}{l}{Position}           & \multicolumn{1}{l|}{Color}             & \multicolumn{1}{l}{}                   \\ \hline
                               & Regular  & 180                                  & 73.3                                   & 76.7                                   & 108.3                                  & 438.3                                  \\
                               & ICD      & 180                                  & \cellcolor{rank2}80             & \cellcolor{rank1}\textbf{80}    & \cellcolor{rank2}130.3          & \cellcolor{rank2}470.3          \\
\multirow{-2}{*}{InstructBLIP} & VCD      & \cellcolor{rank1}\textbf{190} & 65                                     & 58.3                                   & 130                                    & 443.3                                  \\
                               & IFCD (Ours)    & \cellcolor{rank2}185          & \cellcolor{rank1}\textbf{85}    & \cellcolor{rank2}63.3           & \cellcolor{rank1}\textbf{138.3} & \cellcolor{rank1}\textbf{471.6} \\ \cline{2-7} 
                               & Regular  & 180                                  & 123.3                                  & 105                                    & 158.3                                  & 566.6                                  \\
                               & ICD      & 180                                  & 123.3                                  & 110                                    & 158.3                                  & 571.6                                  \\
\multirow{-2}{*}{LLaVA 1.5}    & VCD      & 180                                  & \cellcolor{rank2}125            & \cellcolor{rank2}115            & 153.3                                  & \cellcolor{rank2}573.3          \\
                               & IFCD (Ours)    & \cellcolor{rank1}\textbf{185} & \cellcolor{rank1}\textbf{163.3} & \cellcolor{rank1}\textbf{133.3} & \cellcolor{rank1}\textbf{163.3} & \cellcolor{rank1}\textbf{644.9} \\ \hline
\end{tabular}
}
\end{sc}
\end{small}
\end{center}
\label{tab:subset_mme_blip}
\end{table}

\textbf{Results on MSCOCO~}
We select MSCOCO for the text generation task due to its diverse, multidomain content, which better reflects LVLMs' susceptibility to object hallucination. This provides a distinct evaluation from the yes/no tasks of POPE and MME. Table \ref{tab:caption} compares IFCD with baseline methods on the image caption generation task with the prompt: ``\texttt{Please describe this image in detail.}''. IFCD significantly reduces the proportion of hallucinated objects and sentences, with average reductions of 4.5\% and 21.6\% compared with direct decoding, respectively. Meanwhile, the quality of text generation in the LVLM remains at an average level, indicating that IFCD effectively balances hallucination mitigation with maintaining text generation quality.

\begin{figure*}
\begin{center}
\subfigure[InstructBLIP]{   \includegraphics[width=0.5\linewidth]{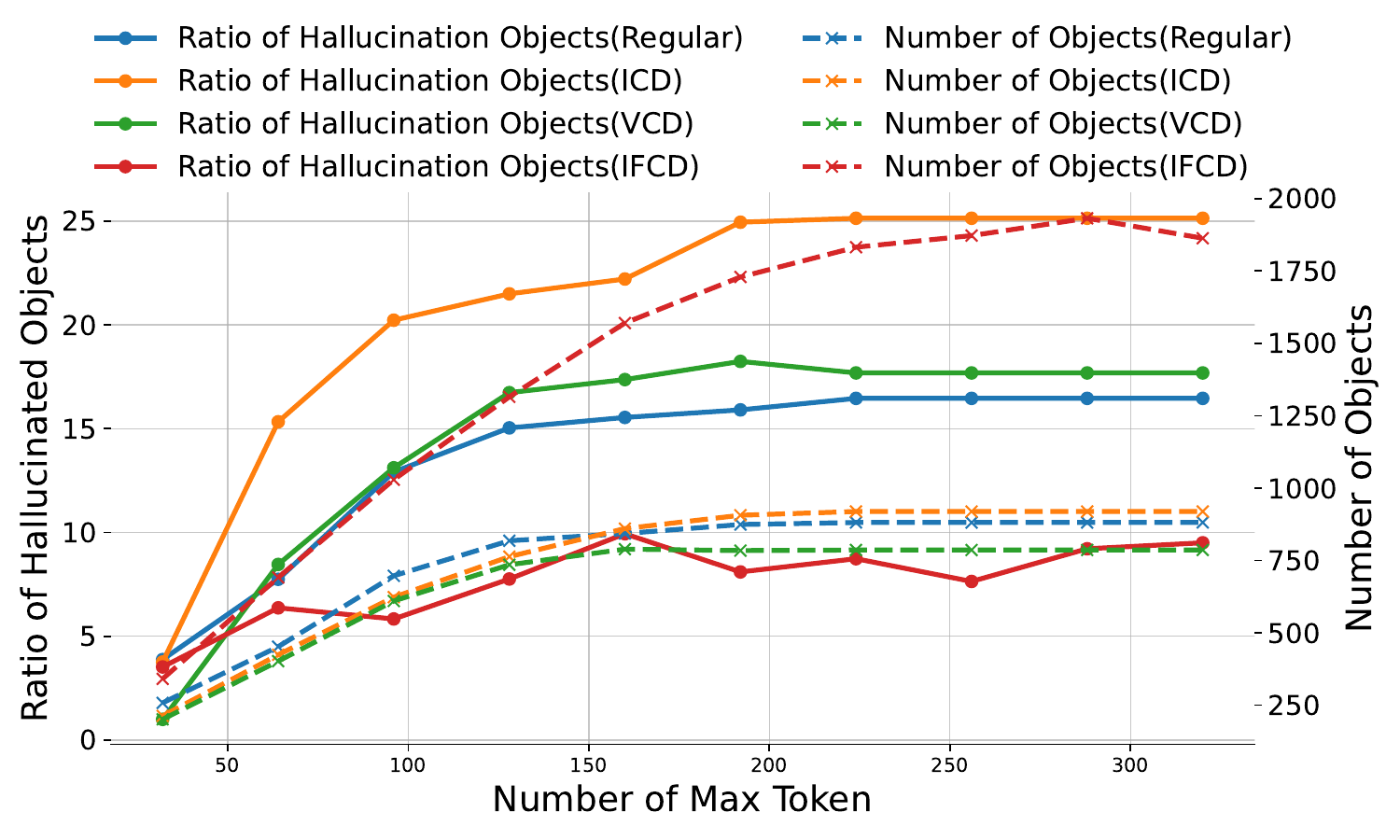}}\subfigure[LLaVA 1.5]{
\includegraphics[width=0.5\linewidth]{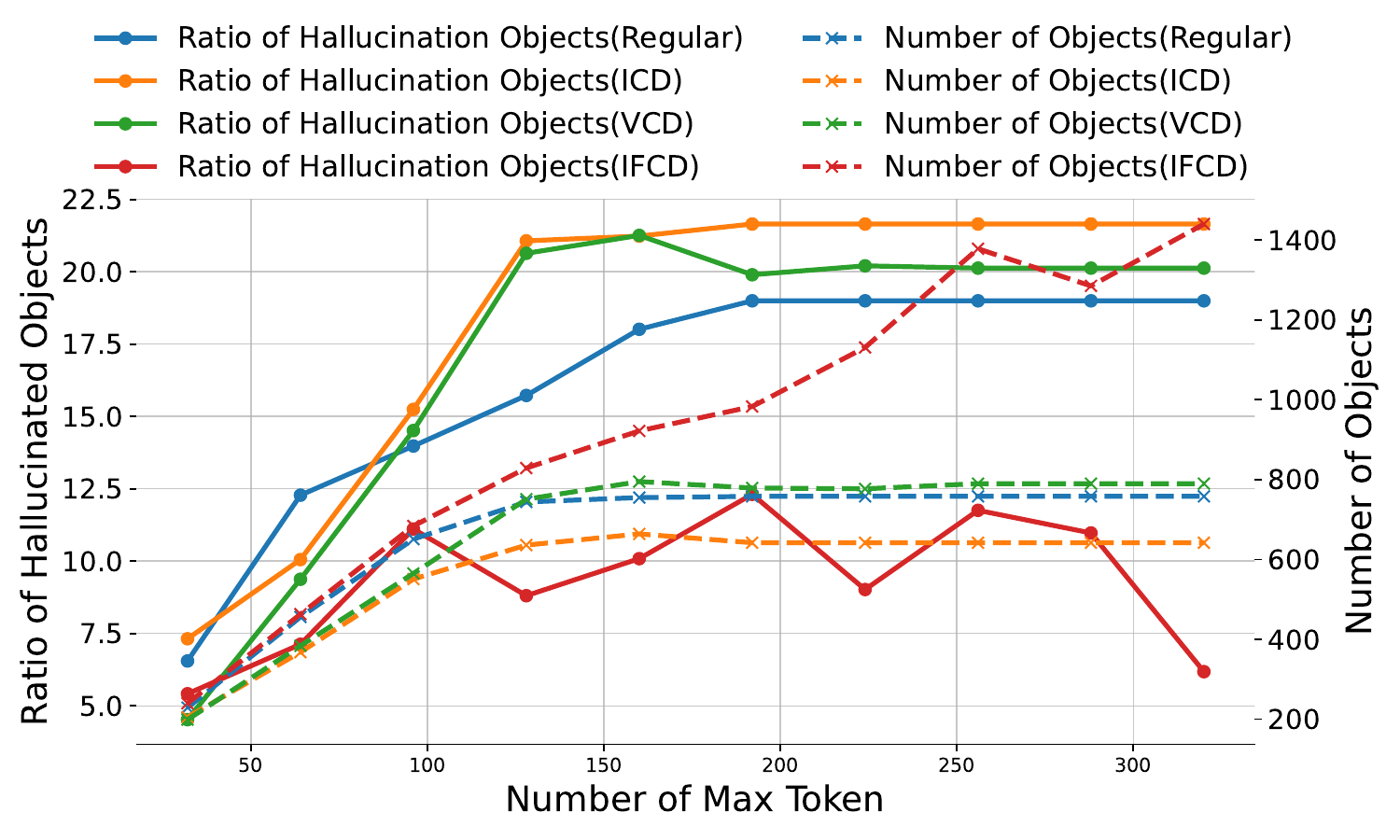}
}
\vskip -0.1in
\caption{Comparison IFCD and regular decoding on the ratio of hallucination objects ($\text{CHAIR}_i$) with respect to the number of max tokens. IFCD maintains a low ratio of hallucination objects while increasing the number of objects.}
\label{fig:ablation_max_token}
\end{center}
\vskip -0.2in
\end{figure*}

\begin{table}[h]
\vskip -0.1in
\caption{Results of InstructBLIP and LLaVA 1.5 on MSCOCO to demonstrate the capacity of IFCD in long-form text generation. \textbf{Bold} and \textcolor{rank1}{\textbf{orange}} indicate the best resutls.}
\begin{center}
\begin{small}
\begin{sc}
\resizebox{0.48\textwidth}{!}{
\begin{tabular}{llccc}
\hline
Model                          & Method      & \multicolumn{1}{l}{$\text{CHAIR}_s$↓}            & \multicolumn{1}{l}{$\text{CHAIR}_i$↓}            & \multicolumn{1}{l}{BLEU↑}              \\ \hline
                               & Regular     & 48                                    & 13.9                                  & \cellcolor{rank1}\textbf{9.8}  \\
                               & ICD         & 58.2                                  & 18.5                                  & 8.4                                   \\
\multirow{-2}{*}{InstructBLIP} & VCD         & 54.8                                  & 16.2                                  & 9.1                                   \\
                               & IFCD (Ours) & \cellcolor{rank1}\textbf{39.6} & \cellcolor{rank1}\textbf{11.2} & 8.4                                   \\ \cline{2-5} 
                               & Regular     & 20                                    & 15.2                                  & 9                                     \\
                               & ICD         & 50.8                                  & 16.9                                  & 8.5                                   \\
\multirow{-2}{*}{LLaVA 1.5}    & VCD         & 21.8                                  & 11                                    & \cellcolor{rank1}\textbf{10.8} \\
                               & IFCD (Ours) & \cellcolor{rank1}\textbf{13.2} & \cellcolor{rank1}\textbf{5.6}  & 8                                     \\ \hline
\end{tabular}
}
\end{sc}
\end{small}
\end{center}
\label{tab:caption}
\end{table}

Furthermore, we also investigate the performance of IFCD  under different maximum generation length settings. Figure \ref{fig:ablation_max_token} illustrates the number of objects generated (dashed line) and the ratio of hallucinated objects (solid line) for 100 randomly selected images from the MSCOCO 2014 validation split. This experiment provides a comprehensive evaluation of IFCD's robustness. Intuitively, an increase in the number of generated objects would typically correspond to a parallel increase in the hallucinatory objects. However, IFCD effectively maintains a low level of object hallucination ratios, even as the number of generated objects continues to grow. These results highlight the generalizability and effectiveness of IFCD in diverse task types, including truthfulness classification and text generation.

\textbf{Case Study on LLaVA-Bench   } \label{llava-bench}
We conduct a qualitative experiment on LLaVA-Bench to demonstrate the performance of IFCD directly. The results of LLaVA-Bench are shown in Appendix \ref{llava_bench_apex} due to the space limits. In each case, we compare baseline methods and IFCD. Then the hallucination contents are marked by red. It is worth noting that IFCD provides rich information while reducing the ratio of hallucinated objects, confirming its robustness. 

\section{Analysis and Ablation Studies} \label{sec: analyse}
\textbf{Ablation Study  } 
The core of implementing IFCD lies in identifying the two token distributions employed for contrast, which must have a gap in the hallucination level. This enables the contrastive decoding to effectively subtract the high hallucination level portion of the distribution, thereby mitigating the object hallucinations in the final token distribution. In IFCD, we designate the distribution that undergoes anti-hallucinations as $P^+$ and the distribution outputted by hallucination-inducing as $P^-$. To validate the effectiveness and stability of IFCD, we conduct ablation experiments that comprehensively compare the results generated by various combinations of these distributions.

In Table \ref{tab: ab component}, we conduct the ablation study on the combinations of distributions, showing the effectiveness and robustness of IFCD. Specifically, contrastive decoding with negative editing and original distribution is a competitive method, which could be attributed to the effect of hallucination-inducing from internal representation editing. In addition, since TruthX could be used to mitigate object hallucinations alone, IFCD overperforms it by a wide margin, manifesting the effectiveness of IFCD. Among all LVLMs, IFCD consistently demonstrates effectiveness and robustness in maintaining a low ratio of hallucination.

\begin{table}[h]
\caption{Ablation of IFCD components. “EDITING”: decoding with positive editing without contrastive decoding; ``w/o NEG'': contrastive decoding with positive editing and original distribution; ``w/o POS'': contrastive decoding with negative editing and original distribution. The best performance is marked by \textbf{bold} and \textbf{\textcolor{rank1}{orange}}, and the second one is marked by \textcolor{rank2}{cyan}.}
\begin{center}
\begin{small}
\begin{sc}
\resizebox{0.48\textwidth}{!}{
\begin{tabular}{llcc}
\hline
Model & Method      & $\text{CHAIR}_s$↓                               & $\text{CHAIR}_i$↓                              \\ \hline
      & Editing     & 57                                    & 15                                   \\
InstructBLIP  & IFCD w/o neg & 71                                    & 19.9                                 \\
      & IFCD w/o pos & \cellcolor{rank1}\textbf{28}   & \cellcolor{rank1}\textbf{7.6} \\ \cline{2-4} 
      & IFCD (ours)        & \cellcolor{rank2}39.6          & \cellcolor{rank2}11.2         \\ \hline 
      & Editing     & \cellcolor{rank2}27            & \cellcolor{rank2}9.4          \\
LLaVA 1.5 & IFCD w/o neg & 44                                    & 11.8                                 \\
      & IFCD w/o pos & 46                                    & 14                                   \\ \cline{2-4} 
      & IFCD (ours)        & \cellcolor{rank1}\textbf{13.2} & \cellcolor{rank1}\textbf{5.6} \\ \hline
\end{tabular}}
\end{sc}
\end{small}
\end{center}
\label{tab: ab component}
\end{table}

\textbf{Internal Representation Editing } 
The initial step of IFCD involves internal representation editing, making editing strength and the number of editing layers critical hyperparameters. Thus, we examine the impact of varying editing strength and the number of editing layers on IFCD performance with metrics $\text{CHAIR}_s$ and $\text{CHAIR}_i$.

\begin{figure}[h!]
\begin{center}
    \includegraphics[width=1\linewidth]{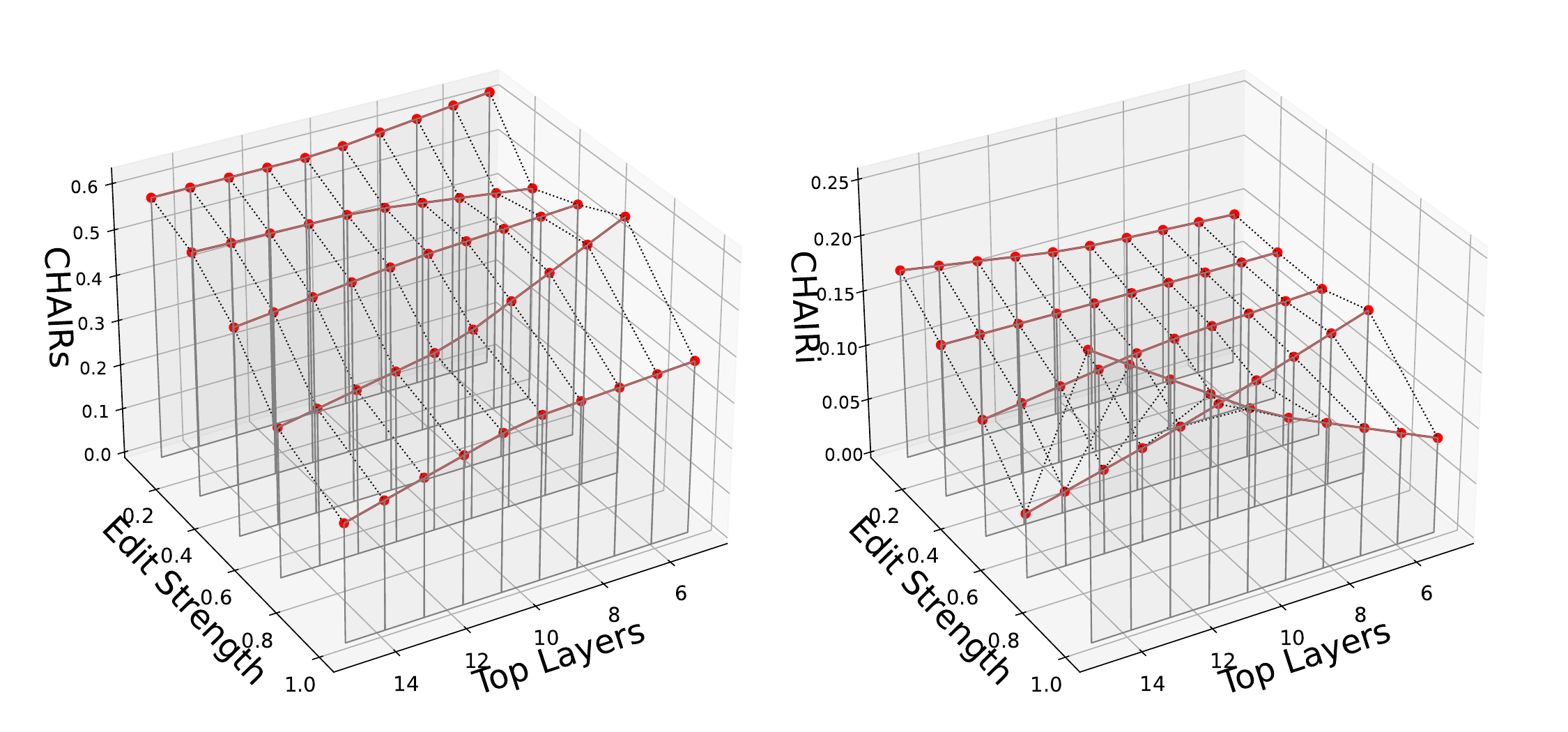}
    \vskip -0.1in
    \caption{CHAIR scores vary with editing strength and layers.}
    \label{fig:edit_param}
\end{center}
\end{figure}

Figure \ref{fig:edit_param} presents the impact of editing strength and layers when editing the internal representation on the effectiveness of IFCD in hallucinations mitigation. For the $\text{CHAIR}_s$ metric, the results demonstrate continuous optimization as both the editing strength and the number of editing layers increase. However, regarding the $\text{CHAIR}_i$ metric, performance degradation occurs when the editing strength reaches 1. Furthermore, the relationship between $\text{CHAIR}_i$ and the number of editing layers becomes inversely correlated. This observation strongly suggests that an editing strength of 1 is a critical threshold, beyond which further adjustments to the internal representation yield diminishing performance.

\textbf{Contrastive Decoding Strength   }
After internal representation editing, contrastive decoding is employed to recalibrate distribution. During this stage, the key parameter is contrastive decoding strength $\alpha$, which controls the extent of contrastive decoding. Intuitively, when the gap of distributions involved in contrast is small, larger $\alpha$ is needed, and vice versa.

\begin{figure}[h]
\centering
\includegraphics[width=0.9\linewidth]{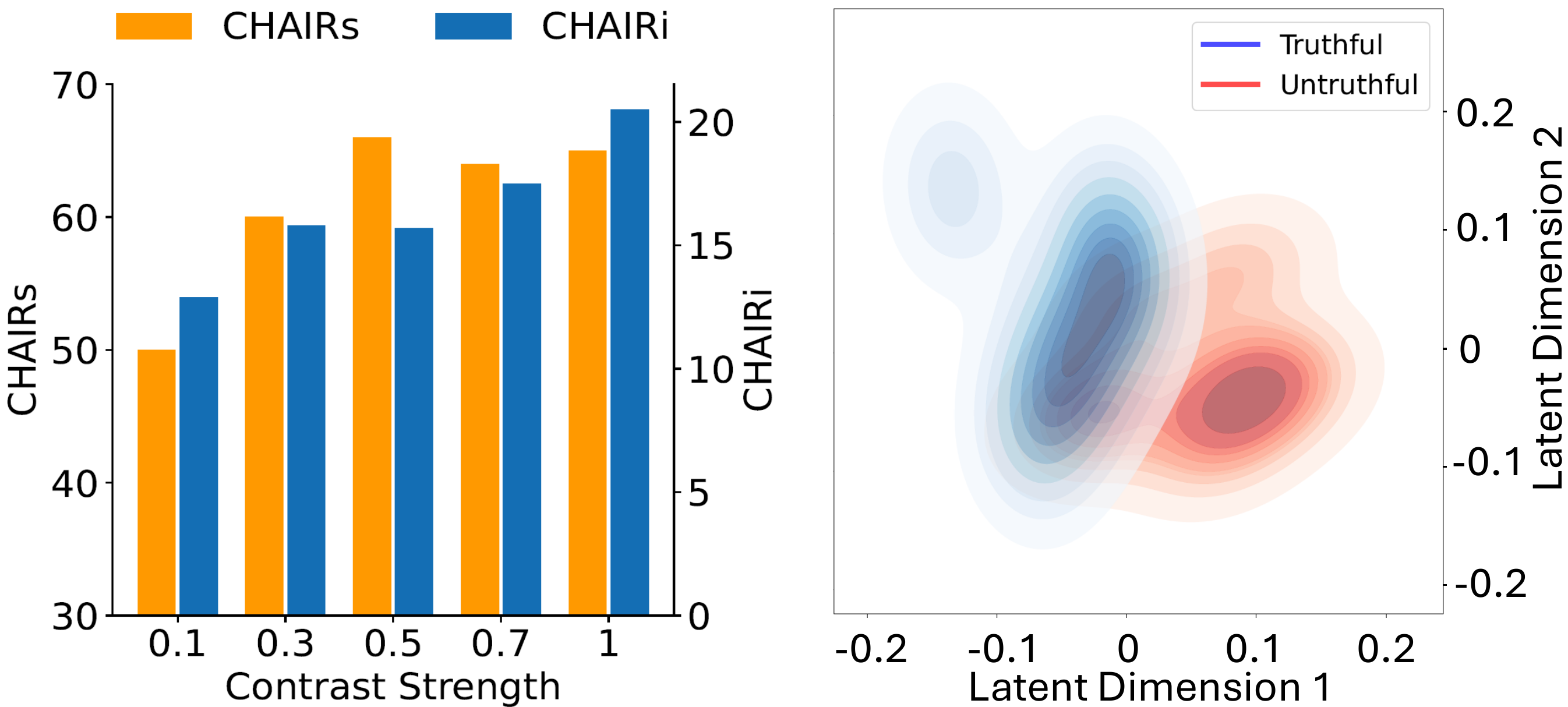}
\caption{IFCD performance with different contrast strengths and the capacity of identifying truthfulness. The order of magnitude of the PCA figure is 1e-7.}
\label{fig: last ablation}
\end{figure}

As shown in the left part of Figure \ref{fig: last ablation}, the small $\alpha$ leads top performance, denoting the gap of distributions involved in contrastive decoding is striking, and editing internal representation is a promising way to expose LVLMs' hallucinations preference.

\textbf{The Capacity of Editing Internal Representation } To investigate the effect of editing internal representations, we explore the latent space of $\mathrm{TruthEnc(\cdot)}$, which maps and edits the internal representation of LVLMs in latent space. We provide responses that differ in truthfulness to the LVLMs and map internal representations with different truthfulness generated during the inference process into the latent space of $\mathrm{TruthEnc(\cdot)}$. We then apply Principal Component Analysis (PCA) to reduce the dimensionality of the latent space to two dimensions to visualize the latent space as illustrated in the right part of Figure \ref{fig: last ablation}. The figure clearly illustrates a distinct separation between internal representations that differ in truthfulness, demonstrating the model's ability to effectively identify and modify internal representations. This observation also suggests that only a minimal contrastive decoding strength is required to achieve optimal performance.
\section{Conclusion}
In this paper, we propose a novel method, IFCD, to mitigate object hallucination. 
In this method, we apply contrastive decoding based on truthfulness editing of internal representations to eliminate hallucinatory elements that are actively induced and closely aligned with statistic biases.
Extensive experimentation across diverse benchmarks and LVLMs confirms the efficacy of IFCD.

\bibliography{ref}

\begin{thebibliography}{48}
\providecommand{\natexlab}[1]{#1}
\providecommand{\url}[1]{\texttt{#1}}
\expandafter\ifx\csname urlstyle\endcsname\relax
  \providecommand{\doi}[1]{doi: #1}\else
  \providecommand{\doi}{doi: \begingroup \urlstyle{rm}\Url}\fi

\bibitem[Achiam et~al.(2023)Achiam, Adler, Agarwal, Ahmad, Akkaya, Aleman, Almeida, Altenschmidt, Altman, Anadkat, et~al.]{achiam2023gpt}
Achiam, J., Adler, S., Agarwal, S., Ahmad, L., Akkaya, I., Aleman, F.~L., Almeida, D., Altenschmidt, J., Altman, S., Anadkat, S., et~al.
\newblock Gpt-4 technical report.
\newblock \emph{arXiv preprint arXiv:2303.08774}, 2023.

\bibitem[Asai et~al.(2023)Asai, Wu, Wang, Sil, and Hajishirzi]{asai2023self}
Asai, A., Wu, Z., Wang, Y., Sil, A., and Hajishirzi, H.
\newblock Self-rag: Self-reflective retrieval augmented generation.
\newblock In \emph{NeurIPS 2023 Workshop on Instruction Tuning and Instruction Following}, 2023.

\bibitem[Bai et~al.(2023)Bai, Bai, Yang, Wang, Tan, Wang, Lin, Zhou, and Zhou]{bai2023qwen}
Bai, J., Bai, S., Yang, S., Wang, S., Tan, S., Wang, P., Lin, J., Zhou, C., and Zhou, J.
\newblock Qwen-vl: A frontier large vision-language model with versatile abilities.
\newblock \emph{arXiv preprint arXiv:2308.12966}, 2023.

\bibitem[Biten et~al.(2022)Biten, G{\'o}mez, and Karatzas]{biten2022let}
Biten, A.~F., G{\'o}mez, L., and Karatzas, D.
\newblock Let there be a clock on the beach: Reducing object hallucination in image captioning.
\newblock In \emph{Proceedings of the IEEE/CVF Winter Conference on Applications of Computer Vision}, pp.\  1381--1390, 2022.

\bibitem[Chen et~al.(2024{\natexlab{a}})Chen, Xiong, Liu, Wu, Xiao, Gao, and He]{chen2024context}
Chen, S., Xiong, M., Liu, J., Wu, Z., Xiao, T., Gao, S., and He, J.
\newblock In-context sharpness as alerts: An inner representation perspective for hallucination mitigation.
\newblock In \emph{ICLR 2024 Workshop on Reliable and Responsible Foundation Models}, 2024{\natexlab{a}}.

\bibitem[Chen et~al.(2024{\natexlab{b}})Chen, Xiong, Liu, Wu, Xiao, Gao, and He]{chen2024incontext}
Chen, S., Xiong, M., Liu, J., Wu, Z., Xiao, T., Gao, S., and He, J.
\newblock In-context sharpness as alerts: An inner representation perspective for hallucination mitigation.
\newblock In \emph{ICLR 2024 Workshop on Reliable and Responsible Foundation Models}, 2024{\natexlab{b}}.
\newblock URL \url{https://openreview.net/forum?id=24U6vAHnYM}.

\bibitem[Chen et~al.(2024{\natexlab{c}})Chen, Zhao, Luo, Yao, Li, and Zhou]{chenhalc}
Chen, Z., Zhao, Z., Luo, H., Yao, H., Li, B., and Zhou, J.
\newblock Halc: Object hallucination reduction via adaptive focal-contrast decoding.
\newblock In \emph{Forty-first International Conference on Machine Learning}, 2024{\natexlab{c}}.

\bibitem[Chiang et~al.(2023)Chiang, Li, Lin, Sheng, Wu, Zhang, Zheng, Zhuang, Zhuang, Gonzalez, et~al.]{chiang2023vicuna}
Chiang, W.-L., Li, Z., Lin, Z., Sheng, Y., Wu, Z., Zhang, H., Zheng, L., Zhuang, S., Zhuang, Y., Gonzalez, J.~E., et~al.
\newblock Vicuna: An open-source chatbot impressing gpt-4 with 90\%* chatgpt quality.
\newblock \emph{See https://vicuna. lmsys. org (accessed 14 April 2023)}, 2\penalty0 (3):\penalty0 6, 2023.

\bibitem[Dhuliawala et~al.(2024)Dhuliawala, Komeili, Xu, Raileanu, Li, Celikyilmaz, and Weston]{dhuliawala-etal-2024-chain}
Dhuliawala, S., Komeili, M., Xu, J., Raileanu, R., Li, X., Celikyilmaz, A., and Weston, J.
\newblock Chain-of-verification reduces hallucination in large language models.
\newblock In Ku, L.-W., Martins, A., and Srikumar, V. (eds.), \emph{Findings of the Association for Computational Linguistics: ACL 2024}, pp.\  3563--3578, Bangkok, Thailand, August 2024. Association for Computational Linguistics.
\newblock \doi{10.18653/v1/2024.findings-acl.212}.
\newblock URL \url{https://aclanthology.org/2024.findings-acl.212/}.

\bibitem[Dosovitskiy et~al.(2021)Dosovitskiy, Beyer, Kolesnikov, Weissenborn, Zhai, Unterthiner, Dehghani, Minderer, Heigold, Gelly, Uszkoreit, and Houlsby]{dosovitskiy2021an}
Dosovitskiy, A., Beyer, L., Kolesnikov, A., Weissenborn, D., Zhai, X., Unterthiner, T., Dehghani, M., Minderer, M., Heigold, G., Gelly, S., Uszkoreit, J., and Houlsby, N.
\newblock An image is worth 16x16 words: Transformers for image recognition at scale.
\newblock In \emph{International Conference on Learning Representations}, 2021.
\newblock URL \url{https://openreview.net/forum?id=YicbFdNTTy}.

\bibitem[Driess et~al.(2023)Driess, Xia, Sajjadi, Lynch, Chowdhery, Ichter, Wahid, Tompson, Vuong, Yu, Huang, Chebotar, Sermanet, Duckworth, Levine, Vanhoucke, Hausman, Toussaint, Greff, Zeng, Mordatch, and Florence]{pmlr-v202-driess23a}
Driess, D., Xia, F., Sajjadi, M. S.~M., Lynch, C., Chowdhery, A., Ichter, B., Wahid, A., Tompson, J., Vuong, Q., Yu, T., Huang, W., Chebotar, Y., Sermanet, P., Duckworth, D., Levine, S., Vanhoucke, V., Hausman, K., Toussaint, M., Greff, K., Zeng, A., Mordatch, I., and Florence, P.
\newblock {P}a{LM}-e: An embodied multimodal language model.
\newblock In Krause, A., Brunskill, E., Cho, K., Engelhardt, B., Sabato, S., and Scarlett, J. (eds.), \emph{Proceedings of the 40th International Conference on Machine Learning}, volume 202 of \emph{Proceedings of Machine Learning Research}, pp.\  8469--8488. PMLR, 23--29 Jul 2023.
\newblock URL \url{https://proceedings.mlr.press/v202/driess23a.html}.

\bibitem[Fu et~al.(2023)Fu, Chen, Shen, Qin, Zhang, Lin, Yang, Zheng, Li, Sun, et~al.]{fu2023mme}
Fu, C., Chen, P., Shen, Y., Qin, Y., Zhang, M., Lin, X., Yang, J., Zheng, X., Li, K., Sun, X., et~al.
\newblock Mme: A comprehensive evaluation benchmark for multimodal large language models.
\newblock \emph{arXiv preprint arXiv:2306.13394}, 2023.

\bibitem[Han et~al.(2024)Han, Kossen, Razzak, Schut, Malik, and Gal]{han2024semantic}
Han, J., Kossen, J., Razzak, M., Schut, L., Malik, S.~A., and Gal, Y.
\newblock Semantic entropy probes: Robust and cheap hallucination detection in llms.
\newblock In \emph{ICML 2024 Workshop on Foundation Models in the Wild}, 2024.

\bibitem[Hu et~al.(2024)Hu, Ru, Qiu, Guo, Zhang, Xu, Luo, Liu, Zhang, and Zhang]{hu-etal-2024-knowledge}
Hu, X., Ru, D., Qiu, L., Guo, Q., Zhang, T., Xu, Y., Luo, Y., Liu, P., Zhang, Y., and Zhang, Z.
\newblock Knowledge-centric hallucination detection.
\newblock In Al-Onaizan, Y., Bansal, M., and Chen, Y.-N. (eds.), \emph{Proceedings of the 2024 Conference on Empirical Methods in Natural Language Processing}, pp.\  6953--6975, Miami, Florida, USA, November 2024. Association for Computational Linguistics.
\newblock \doi{10.18653/v1/2024.emnlp-main.395}.
\newblock URL \url{https://aclanthology.org/2024.emnlp-main.395/}.

\bibitem[Ji et~al.(2023)Ji, Lee, Frieske, Yu, Su, Xu, Ishii, Bang, Madotto, and Fung]{ji2023survey}
Ji, Z., Lee, N., Frieske, R., Yu, T., Su, D., Xu, Y., Ishii, E., Bang, Y.~J., Madotto, A., and Fung, P.
\newblock Survey of hallucination in natural language generation.
\newblock \emph{ACM Computing Surveys}, 55\penalty0 (12):\penalty0 1--38, 2023.

\bibitem[Leng et~al.(2024)Leng, Zhang, Chen, Li, Lu, Miao, and Bing]{leng2024mitigating}
Leng, S., Zhang, H., Chen, G., Li, X., Lu, S., Miao, C., and Bing, L.
\newblock Mitigating object hallucinations in large vision-language models through visual contrastive decoding.
\newblock In \emph{Proceedings of the IEEE/CVF Conference on Computer Vision and Pattern Recognition}, pp.\  13872--13882, 2024.

\bibitem[Li et~al.(2023{\natexlab{a}})Li, Li, Le, Wang, Savarese, and Hoi]{li-etal-2023-lavis}
Li, D., Li, J., Le, H., Wang, G., Savarese, S., and Hoi, S.~C.
\newblock {LAVIS}: A one-stop library for language-vision intelligence.
\newblock In Bollegala, D., Huang, R., and Ritter, A. (eds.), \emph{Proceedings of the 61st Annual Meeting of the Association for Computational Linguistics (Volume 3: System Demonstrations)}, pp.\  31--41, Toronto, Canada, July 2023{\natexlab{a}}. Association for Computational Linguistics.
\newblock \doi{10.18653/v1/2023.acl-demo.3}.
\newblock URL \url{https://aclanthology.org/2023.acl-demo.3/}.

\bibitem[Li et~al.(2023{\natexlab{b}})Li, Li, Savarese, and Hoi]{li2023blip}
Li, J., Li, D., Savarese, S., and Hoi, S.
\newblock Blip-2: Bootstrapping language-image pre-training with frozen image encoders and large language models.
\newblock In \emph{International conference on machine learning}, pp.\  19730--19742. PMLR, 2023{\natexlab{b}}.

\bibitem[Li et~al.(2023{\natexlab{c}})Li, Holtzman, Fried, Liang, Eisner, Hashimoto, Zettlemoyer, and Lewis]{li-etal-2023-contrastive}
Li, X.~L., Holtzman, A., Fried, D., Liang, P., Eisner, J., Hashimoto, T., Zettlemoyer, L., and Lewis, M.
\newblock Contrastive decoding: Open-ended text generation as optimization.
\newblock In Rogers, A., Boyd-Graber, J., and Okazaki, N. (eds.), \emph{Proceedings of the 61st Annual Meeting of the Association for Computational Linguistics (Volume 1: Long Papers)}, pp.\  12286--12312, Toronto, Canada, July 2023{\natexlab{c}}. Association for Computational Linguistics.
\newblock \doi{10.18653/v1/2023.acl-long.687}.
\newblock URL \url{https://aclanthology.org/2023.acl-long.687/}.

\bibitem[Li et~al.(2023{\natexlab{d}})Li, Du, Zhou, Wang, Zhao, and Wen]{li-etal-2023-evaluating}
Li, Y., Du, Y., Zhou, K., Wang, J., Zhao, X., and Wen, J.-R.
\newblock Evaluating object hallucination in large vision-language models.
\newblock In Bouamor, H., Pino, J., and Bali, K. (eds.), \emph{Proceedings of the 2023 Conference on Empirical Methods in Natural Language Processing}, pp.\  292--305, Singapore, December 2023{\natexlab{d}}. Association for Computational Linguistics.
\newblock \doi{10.18653/v1/2023.emnlp-main.20}.
\newblock URL \url{https://aclanthology.org/2023.emnlp-main.20/}.

\bibitem[Lin et~al.(2014)Lin, Maire, Belongie, Hays, Perona, Ramanan, Doll{\'a}r, and Zitnick]{MSCOCO}
Lin, T.-Y., Maire, M., Belongie, S., Hays, J., Perona, P., Ramanan, D., Doll{\'a}r, P., and Zitnick, C.~L.
\newblock Microsoft coco: Common objects in context.
\newblock In Fleet, D., Pajdla, T., Schiele, B., and Tuytelaars, T. (eds.), \emph{Computer Vision -- ECCV 2014}, pp.\  740--755, Cham, 2014. Springer International Publishing.
\newblock ISBN 978-3-319-10602-1.

\bibitem[Liu et~al.(2024)Liu, Li, Wu, and Lee]{liu2024visual}
Liu, H., Li, C., Wu, Q., and Lee, Y.~J.
\newblock Visual instruction tuning.
\newblock \emph{Advances in neural information processing systems}, 36, 2024.

\bibitem[Mahaut et~al.(2024)Mahaut, Aina, Czarnowska, Hardalov, M{\"u}ller, and Marquez]{mahaut-etal-2024-factual}
Mahaut, M., Aina, L., Czarnowska, P., Hardalov, M., M{\"u}ller, T., and Marquez, L.
\newblock Factual confidence of {LLM}s: on reliability and robustness of current estimators.
\newblock In Ku, L.-W., Martins, A., and Srikumar, V. (eds.), \emph{Proceedings of the 62nd Annual Meeting of the Association for Computational Linguistics (Volume 1: Long Papers)}, pp.\  4554--4570, Bangkok, Thailand, August 2024. Association for Computational Linguistics.
\newblock \doi{10.18653/v1/2024.acl-long.250}.
\newblock URL \url{https://aclanthology.org/2024.acl-long.250/}.

\bibitem[Min et~al.(2024)Min, Kim, Lee, Lee, and Jung]{min2024mitigating}
Min, K., Kim, M., Lee, K.-i., Lee, D., and Jung, K.
\newblock Mitigating hallucinations in lvlms via summary-guided decoding.
\newblock In \emph{Neurips Safe Generative AI Workshop 2024}, 2024.

\bibitem[Pal et~al.(2023)Pal, Umapathi, and Sankarasubbu]{pal-etal-2023-med}
Pal, A., Umapathi, L.~K., and Sankarasubbu, M.
\newblock {M}ed-{HALT}: Medical domain hallucination test for large language models.
\newblock In Jiang, J., Reitter, D., and Deng, S. (eds.), \emph{Proceedings of the 27th Conference on Computational Natural Language Learning (CoNLL)}, pp.\  314--334, Singapore, December 2023. Association for Computational Linguistics.
\newblock \doi{10.18653/v1/2023.conll-1.21}.
\newblock URL \url{https://aclanthology.org/2023.conll-1.21/}.

\bibitem[Pan et~al.(2024)Pan, Fan, Li, Yu, Fei, Tang, Hong, Zhang, and Sun]{pan2024towards}
Pan, K., Fan, Z., Li, J., Yu, Q., Fei, H., Tang, S., Hong, R., Zhang, H., and Sun, Q.
\newblock Towards unified multimodal editing with enhanced knowledge collaboration.
\newblock In \emph{The Thirty-eighth Annual Conference on Neural Information Processing Systems}, 2024.
\newblock URL \url{https://openreview.net/forum?id=kf80ZS3fVy}.

\bibitem[Papineni et~al.(2002)Papineni, Roukos, Ward, and Zhu]{papineni-etal-2002-bleu}
Papineni, K., Roukos, S., Ward, T., and Zhu, W.-J.
\newblock {B}leu: a method for automatic evaluation of machine translation.
\newblock In Isabelle, P., Charniak, E., and Lin, D. (eds.), \emph{Proceedings of the 40th Annual Meeting of the Association for Computational Linguistics}, pp.\  311--318, Philadelphia, Pennsylvania, USA, July 2002. Association for Computational Linguistics.
\newblock \doi{10.3115/1073083.1073135}.
\newblock URL \url{https://aclanthology.org/P02-1040/}.

\bibitem[Radford et~al.(2021)Radford, Kim, Hallacy, Ramesh, Goh, Agarwal, Sastry, Askell, Mishkin, Clark, et~al.]{radford2021learning}
Radford, A., Kim, J.~W., Hallacy, C., Ramesh, A., Goh, G., Agarwal, S., Sastry, G., Askell, A., Mishkin, P., Clark, J., et~al.
\newblock Learning transferable visual models from natural language supervision.
\newblock In \emph{International conference on machine learning}, pp.\  8748--8763. PMLR, 2021.

\bibitem[Rafailov et~al.(2024)Rafailov, Sharma, Mitchell, Manning, Ermon, and Finn]{rafailov2024direct}
Rafailov, R., Sharma, A., Mitchell, E., Manning, C.~D., Ermon, S., and Finn, C.
\newblock Direct preference optimization: Your language model is secretly a reward model.
\newblock \emph{Advances in Neural Information Processing Systems}, 36, 2024.

\bibitem[Rohrbach et~al.(2018)Rohrbach, Hendricks, Burns, Darrell, and Saenko]{rohrbach-etal-2018-object}
Rohrbach, A., Hendricks, L.~A., Burns, K., Darrell, T., and Saenko, K.
\newblock Object hallucination in image captioning.
\newblock In Riloff, E., Chiang, D., Hockenmaier, J., and Tsujii, J. (eds.), \emph{Proceedings of the 2018 Conference on Empirical Methods in Natural Language Processing}, pp.\  4035--4045, Brussels, Belgium, October-November 2018. Association for Computational Linguistics.
\newblock \doi{10.18653/v1/D18-1437}.
\newblock URL \url{https://aclanthology.org/D18-1437/}.

\bibitem[Sahoo et~al.(2024)Sahoo, Meharia, Ghosh, Saha, Jain, and Chadha]{sahoo2024comprehensive}
Sahoo, P., Meharia, P., Ghosh, A., Saha, S., Jain, V., and Chadha, A.
\newblock A comprehensive survey of hallucination in large language, image, video and audio foundation models.
\newblock \emph{Findings of the Association for Computational Linguistics: EMNLP 2024}, pp.\  11709--11724, 2024.

\bibitem[Sanderson(2023)]{sanderson2023gpt}
Sanderson, K.
\newblock Gpt-4 is here: what scientists think.
\newblock \emph{Nature}, 615\penalty0 (7954):\penalty0 773, 2023.

\bibitem[Singhal et~al.(2024)Singhal, Law, Kassner, Gupta, Duan, Damle, and Li]{singhal2024multilingual}
Singhal, A., Law, T., Kassner, C., Gupta, A., Duan, E., Damle, A., and Li, R.
\newblock Multilingual fact-checking using llms.
\newblock In \emph{Proceedings of the Third Workshop on NLP for Positive Impact}, pp.\  13--31, 2024.

\bibitem[Song \& Huang(2024)Song and Huang]{song2024hscl}
Song, Z. and Huang, S.
\newblock Hscl-rl: Mitigating hallucinations in multimodal large language models.
\newblock In \emph{NeurIPS 2024 Workshop on Open-World Agents}, 2024.

\bibitem[Stiennon et~al.(2020)Stiennon, Ouyang, Wu, Ziegler, Lowe, Voss, Radford, Amodei, and Christiano]{stiennon2020learning}
Stiennon, N., Ouyang, L., Wu, J., Ziegler, D., Lowe, R., Voss, C., Radford, A., Amodei, D., and Christiano, P.~F.
\newblock Learning to summarize with human feedback.
\newblock \emph{Advances in Neural Information Processing Systems}, 33:\penalty0 3008--3021, 2020.

\bibitem[Wang et~al.(2024)Wang, Pan, Ding, and Biemann]{wang-etal-2024-mitigating}
Wang, X., Pan, J., Ding, L., and Biemann, C.
\newblock Mitigating hallucinations in large vision-language models with instruction contrastive decoding.
\newblock In Ku, L.-W., Martins, A., and Srikumar, V. (eds.), \emph{Findings of the Association for Computational Linguistics: ACL 2024}, pp.\  15840--15853, Bangkok, Thailand, August 2024. Association for Computational Linguistics.
\newblock \doi{10.18653/v1/2024.findings-acl.937}.
\newblock URL \url{https://aclanthology.org/2024.findings-acl.937/}.

\bibitem[Wei et~al.(2022)Wei, Bosma, Zhao, Guu, Yu, Lester, Du, Dai, and Le]{wei2022finetuned}
Wei, J., Bosma, M., Zhao, V., Guu, K., Yu, A.~W., Lester, B., Du, N., Dai, A.~M., and Le, Q.~V.
\newblock Finetuned language models are zero-shot learners.
\newblock In \emph{International Conference on Learning Representations}, 2022.
\newblock URL \url{https://openreview.net/forum?id=gEZrGCozdqR}.

\bibitem[Wen et~al.(2024)Wen, Yang, Fu, Wang, Cai, Li, Tao, Li, Linran, Shang, et~al.]{wen2024road}
Wen, L., Yang, X., Fu, D., Wang, X., Cai, P., Li, X., Tao, M., Li, Y., Linran, X., Shang, D., et~al.
\newblock On the road with gpt-4v (ision): Explorations of utilizing visual-language model as autonomous driving agent.
\newblock In \emph{ICLR 2024 Workshop on Large Language Model (LLM) Agents}, 2024.

\bibitem[Wolf et~al.(2020)Wolf, Debut, Sanh, Chaumond, Delangue, Moi, Cistac, Rault, Louf, Funtowicz, et~al.]{wolf2020transformers}
Wolf, T., Debut, L., Sanh, V., Chaumond, J., Delangue, C., Moi, A., Cistac, P., Rault, T., Louf, R., Funtowicz, M., et~al.
\newblock Transformers: State-of-the-art natural language processing.
\newblock \emph{EMNLP 2020}, pp.\ ~38, 2020.

\bibitem[Yang et~al.(2023)Yang, Li, Wang, Lin, Azarnasab, Ahmed, Liu, Liu, Zeng, and Wang]{yang2023mm}
Yang, Z., Li, L., Wang, J., Lin, K., Azarnasab, E., Ahmed, F., Liu, Z., Liu, C., Zeng, M., and Wang, L.
\newblock Mm-react: Prompting chatgpt for multimodal reasoning and action.
\newblock \emph{arXiv preprint arXiv:2303.11381}, 2023.

\bibitem[Ye et~al.(2024{\natexlab{a}})Ye, Xu, Ye, Yan, Hu, Liu, Qian, Zhang, and Huang]{Ye_2024_CVPR}
Ye, Q., Xu, H., Ye, J., Yan, M., Hu, A., Liu, H., Qian, Q., Zhang, J., and Huang, F.
\newblock mplug-owl2: Revolutionizing multi-modal large language model with modality collaboration.
\newblock In \emph{Proceedings of the IEEE/CVF Conference on Computer Vision and Pattern Recognition (CVPR)}, pp.\  13040--13051, June 2024{\natexlab{a}}.

\bibitem[Ye et~al.(2024{\natexlab{b}})Ye, Xu, Ye, Yan, Hu, Liu, Qian, Zhang, and Huang]{ye2024mplug}
Ye, Q., Xu, H., Ye, J., Yan, M., Hu, A., Liu, H., Qian, Q., Zhang, J., and Huang, F.
\newblock mplug-owl2: Revolutionizing multi-modal large language model with modality collaboration.
\newblock In \emph{Proceedings of the IEEE/CVF Conference on Computer Vision and Pattern Recognition}, pp.\  13040--13051, 2024{\natexlab{b}}.

\bibitem[Zhang et~al.(2024)Zhang, Yu, and Feng]{truthx}
Zhang, S., Yu, T., and Feng, Y.
\newblock Truthx: Alleviating hallucinations by editing large language models in truthful space.
\newblock In \emph{Proceedings of the 62th Annual Meeting of the Association for Computational Linguistics (Volume 1: Long Papers)}. Association for Computational Linguistics, 2024.
\newblock URL \url{https://arxiv.org/abs/2402.17811}.

\bibitem[Zhang et~al.(2023)Zhang, Li, Cui, Cai, Liu, Fu, Huang, Zhao, Zhang, Chen, et~al.]{zhang2023siren}
Zhang, Y., Li, Y., Cui, L., Cai, D., Liu, L., Fu, T., Huang, X., Zhao, E., Zhang, Y., Chen, Y., et~al.
\newblock Siren's song in the ai ocean: a survey on hallucination in large language models.
\newblock \emph{arXiv preprint arXiv:2309.01219}, 2023.

\bibitem[Zhao et~al.(2024)Zhao, Deng, Zhang, and Gu]{zhao2024mitigating}
Zhao, L., Deng, Y., Zhang, W., and Gu, Q.
\newblock Mitigating object hallucination in large vision-language models via image-grounded guidance.
\newblock In \emph{Neurips Safe Generative AI Workshop 2024}, 2024.

\bibitem[Zhao et~al.(2023)Zhao, Zhou, Li, Tang, Wang, Hou, Min, Zhang, Zhang, Dong, et~al.]{zhao2023survey}
Zhao, W.~X., Zhou, K., Li, J., Tang, T., Wang, X., Hou, Y., Min, Y., Zhang, B., Zhang, J., Dong, Z., et~al.
\newblock A survey of large language models.
\newblock \emph{arXiv preprint arXiv:2303.18223}, 2023.

\bibitem[Zhou et~al.(2023)Zhou, Cui, Yoon, Zhang, Deng, Finn, Bansal, and Yao]{zhou2023analyzing}
Zhou, Y., Cui, C., Yoon, J., Zhang, L., Deng, Z., Finn, C., Bansal, M., and Yao, H.
\newblock Analyzing and mitigating object hallucination in large vision-language models.
\newblock In \emph{NeurIPS 2023 Workshop on Instruction Tuning and Instruction Following}, 2023.

\bibitem[Zhu et~al.(2024)Zhu, Chen, Shen, Li, and Elhoseiny]{zhu2024minigpt}
Zhu, D., Chen, J., Shen, X., Li, X., and Elhoseiny, M.
\newblock Mini{GPT}-4: Enhancing vision-language understanding with advanced large language models.
\newblock In \emph{The Twelfth International Conference on Learning Representations}, 2024.
\newblock URL \url{https://openreview.net/forum?id=1tZbq88f27}.

\end{thebibliography}
\bibliographystyle{icml2025}

\newpage
\appendix
\onecolumn
\section{Training Effect of Internal Representation Editing Model} \label{training_size_apex}
Training the TruthX, which assesses the truthfulness alignment of internal representations in LVLMs, is a prerequisite for the IFCD method. We train the TruthX using a uniformly sampled set of image-text pairs from the MSCOCO dataset. The specific configuration contains an image as visual information input and two counterfactual responses. For training for $\mathrm{SemEnc(\cdot)}$ to maintain semantic consistency during internal representation editing, truthful and untruthful responses are composed of as many similar tokens as possible \cite{truthx}.


\begin{table}[h]
\caption{IFCD performance with different training sizes of internal representation editing model.}
\vskip 0.1in
\label{tab:train_size}
\begin{center}
\begin{small}
\begin{sc}
\begin{tabular}{c|ccc}
\hline
\textbf{Training Size} & \textbf{Adversarial Acc} & \textbf{Popular Acc} & \textbf{Random Acc} \\ \hline
100           & 82.1            & 84.7        & 87.4       \\
200           & 76              & 77.3        & 77.6       \\
300           & \cellcolor{rank1}\textbf{83.5}            & \cellcolor{rank1}\textbf{86}          & \cellcolor{rank1}\textbf{86.4}       \\
400           & 82.1            & 83.6        & 85.3       \\
500           & 82.3            & 84.2        & 85.5       \\
600           & 80.4            & 81.8        & 82         \\
700           & 75.8            & 77.6        & 79.2       \\ \hline
\end{tabular}
\end{sc}
\end{small}
\end{center}
\end{table}

The performance of TruthX in assessing the truthfulness alignment of internal representations is anticipated to improve with an increase in the size of the training dataset, thereby enhancing the efficacy of the IFCD method. However, validation results indicate that the performance of IFCD reaches its peak accuracy when the training dataset contains 300 samples. Interestingly, further increases in the training data size beyond this threshold result in a decline in IFCD's performance. This decline may be attributed to potential overfitting or the introduction of noise within the additional data.

We compare the performance of IFCD with varying training sizes on the MSCOCO subset of POPE, utilizing three different POPE sampling strategies. The results remain consistent across all strategies. As shown in the left part of Figure \ref{tab:train_size}, when conducting IFCD on LLaVA 1.5 using TruthX models trained with different amounts of training data, it is visually evident that the overall best performance is achieved when the training data size reaches 300 in POPE. Further increasing the training data, however, leads to a significant decline in performance.

\section{Experiments Details} \label{apex: param}
The overall experiment settings are reported in Table \ref{tab: overall setting}. While regular direct decoding follows this setting in each experiment, baseline method VCD and our proposed IFCD follow specific settings. We use the default code for the implementation of two backbone LVLMs, InstructBLIP and LLaVA 1.5 in HuggingFace Transformers Repository \cite{wolf2020transformers}.

The hyper-parameters settings for IFCD in our experiments in Section \ref{4} is reported in Table \ref{tab: ifcd setting}. Specifically, as we discussed in Section \ref{sec: analyse}, there are three major hyper-parameters that actively adjust the effectiveness of IFCD: \textit{Editing Strength}, \textit{Editing Layers}, and \textit{Contrastive Decoding Strength}.

Regrading the comparison of baseline decoding methods VCD and ICD, we adopt the code, and hyper-parameters in the public repositories and papers. We strictly follow the implementation as reported in the paper to reproduce results as Table \ref{tab: vcd setting} (VCD) and Table \ref{tab: icd setting} (ICD).

\begin{table}[h!]
\caption{Overall Experiment Settings.}
\vskip 0.1in
\label{tab: overall setting}
\begin{center}
\begin{small}
\begin{sc}
\begin{tabular}{l|c}
\hline
\textbf{Parameters}                      & \textbf{Value} \\ \hline
Maximum New Token (POPE)        & 32    \\ \hline
Maximum New Token (MME)         & 32    \\ \hline
Maximum New Token (MSCOCO)      & 128   \\ \hline
Maximum New Token (LLaVA-Bench) & 512   \\ \hline
Temperature                     & 1     \\ \hline
Top-K                           & FALSE \\ \hline
Top-p                           & 1     \\ \hline
\end{tabular}
\end{sc}
\end{small}
\end{center}
\end{table}

\begin{table}[h!]
\caption{IFCD Hyperparameter Settings.}
\label{tab: ifcd setting}
\vskip 0.1in
\begin{center}
\begin{small}
\begin{sc}
\begin{tabular}{l|c}
\hline
\textbf{Parameters}                     & \textbf{Value} \\ \hline
Editing Strength                  & 0.5   \\ \hline
Editing Layers                    & 15    \\ \hline
Contrastive Deconding Strength & 0.1     \\ \hline
Adaptive Plausible Threshold   & 0.1   \\ \hline
\end{tabular}
\end{sc}
\end{small}
\end{center}
\end{table}

\begin{table}[h!]
\caption{VCD Hyperparameter Settings.}
\label{tab: vcd setting}
\vskip 0.1in
\begin{center}
\begin{small}
\begin{sc}
\begin{tabular}{l|c}
\hline
\textbf{Parameters}                            & \textbf{Value} \\ \hline
Noise Step (POPE)                     & 999   \\ \hline
Noise Step (except POPE) & 500   \\ \hline
Contrastive Deconding Strength        & 1     \\ \hline
Adaptive Plausible Threshold          & 0.1   \\ \hline
\end{tabular}
\end{sc}
\end{small}
\end{center}
\end{table}

\begin{table}[h!]
\caption{ICD Hyperparameter Settings.}
\label{tab: icd setting}
\vskip 0.1in
\begin{center}
\begin{small}
\begin{sc}
\begin{tabular}{l|c}
\hline
\textbf{Parameters}            & \textbf{Value}                          \\ \hline
Instruction Dirturbance Prompt & ``You are a confused object detector.'' \\ \hline
Contrastive Decoding Strength  & 1                                       \\ \hline
Adaptive Plausible Threshold   & 0.1                                     \\ \hline
\end{tabular}
\end{sc}
\end{small}
\end{center}
\end{table}

\section{MME Experiment detailed Results} \label{appendix_mme}
In Table \ref{tab: full_perception}, we comprehensively present the performance of two LVLM benchmarks on perception-related tasks within the MME benchmark. The results demonstrate that the baseline models exhibit consistent performance patterns, while the employment of IFCD significantly enhances their overall perception capabilities. This improvement is likely attributed to IFCD's ability to effectively mitigate logits that expose object hallucination, thereby recalibrating the LVLM to prioritize visual information rather than relying on pre-existing biases and priors. In contrast, the position, celebrity, and OCR score of IFCD is at a relatively low level on InstructBLIP, while LLaVA 1.5 with IFCD achieves the highest scores on each task among the three decoding methods, suggesting the comparatively weak ability of specific LVLM in these reasoning tasks.
\begin{table}[h!]
\caption{Results on all MME perception-related tasks. The best performance of each setting is marked by \textbf{bold} and \textcolor{rank1}{orange}. The second is marked by \textcolor{rank2}{cyan}.}
\vskip 0.15in
\begin{center}
\begin{small}
\begin{sc}
\resizebox{\textwidth}{!}{
\begin{tabular}{llcccccccccc|c}
\hline
Model                          & Decoding & \multicolumn{1}{l}{Existence}        & \multicolumn{1}{l}{Count}              & \multicolumn{1}{l}{Position}           & \multicolumn{1}{l}{Color}              & \multicolumn{1}{l}{Posters}            & \multicolumn{1}{l}{Celebrity}          & \multicolumn{1}{l}{Scene}               & \multicolumn{1}{l}{Landmark}            & \multicolumn{1}{l}{Artwork}          & \multicolumn{1}{l|}{OCR}             & \multicolumn{1}{l}{Total Score}          \\ \hline
                               & Regular  & 180                                  & 73.3                                   & \cellcolor{rank2}76.6                                   & 108.3                                  & 123.4                                  & \cellcolor{rank1}\textbf{105.5} & 144.75                                  & 126.25                                  & \cellcolor{rank2}99.25                                & \cellcolor{rank1}\textbf{95}  & 1037.35                                  \\
                               & ICD      & 180                                  & \cellcolor{rank2}80             & \cellcolor{rank1}\textbf{80}    & \cellcolor{rank2}130.3          & 116.6                                  & 97.3                                   & 151                                     & 133                                     & \cellcolor{rank1}\textbf{101}                                  & 70                                   & \cellcolor{rank2}1072.2           \\
\multirow{-2}{*}{InstructBLIP} & VCD      & \cellcolor{rank1}\textbf{190} & 65                                     & 58.3                                   & 130                                    & \cellcolor{rank2}135            & 102.9                                  & \cellcolor{rank2}152.25          & \cellcolor{rank2}143.75          & 87                                   & 65                                   & 1064.2                                   \\
                               & IFCD     & \cellcolor{rank2}185          & \cellcolor{rank1}\textbf{85}    & 63.3           & \cellcolor{rank1}\textbf{138.3} & \cellcolor{rank1}\textbf{144.5} & \cellcolor{rank2}103.5          & \cellcolor{rank1}\textbf{163.5}  & \cellcolor{rank1}\textbf{160.75} & \cellcolor{rank1}\textbf{101}                                  & \cellcolor{rank2}80           & \cellcolor{rank1}\textbf{1144.85} \\ \cline{2-13} 
                               & Regular  & 180                                  & 123.3                                  & 105                                    & 158.3                                  & 115.6                                  & 107.6                                  & 145.5                                   & 127.5                                   & 107.5                                & 107.5                                & 1170.3                                   \\
                               & ICD      & 180                                  & 123.3                                  & 110                                    & 158.3                                  & 116.6                                  & 107.9                                  & \cellcolor{rank2}146.75          & 130.5                                   & 110.5                                & \cellcolor{rank2}115          & 1183.85                                  \\
\multirow{-2}{*}{LLaVA 1.5}    & VCD      & 180                                  & \cellcolor{rank2}125            & \cellcolor{rank2}115            & 153.3                                  & \cellcolor{rank2}117            & \cellcolor{rank2}127            & 146                                     & \cellcolor{rank2}132.5           & 107.5                                & 92.5                                 & \cellcolor{rank2}1203.3           \\
                               & IFCD     & \cellcolor{rank1}\textbf{185} & \cellcolor{rank1}\textbf{163.3} & \cellcolor{rank1}\textbf{133.3} & \cellcolor{rank1}\textbf{163.3} & \cellcolor{rank1}\textbf{129.6} & \cellcolor{rank1}\textbf{136.7} & \cellcolor{rank1}\textbf{154.25} & \cellcolor{rank1}\textbf{166.75} & \cellcolor{rank1}\textbf{125} & \cellcolor{rank1}\textbf{125} & \cellcolor{rank1}\textbf{1357.2}  \\ \hline
\end{tabular}}
\end{sc}
\end{small}
\end{center}
\vskip -0.1in
\label{tab: full_perception}
\end{table}

\section{Experiment Results on LLaVA-Bench} \label{llava_bench_apex}
As discussed in Section \ref{llava-bench}, we leverage LLaVA-Bench as a case study to compare the outputs of IFCD with other methods qualitatively. All methods use the settings as Section \ref{implementation_detail}. In all cases, red fonts indicate object hallucination, including object existence, attribute, or relationship hallucination.
\begin{figure}[h!]
\begin{center}
    \vskip 0.2in
    \subfigure{
    \includegraphics[width=0.8\linewidth]{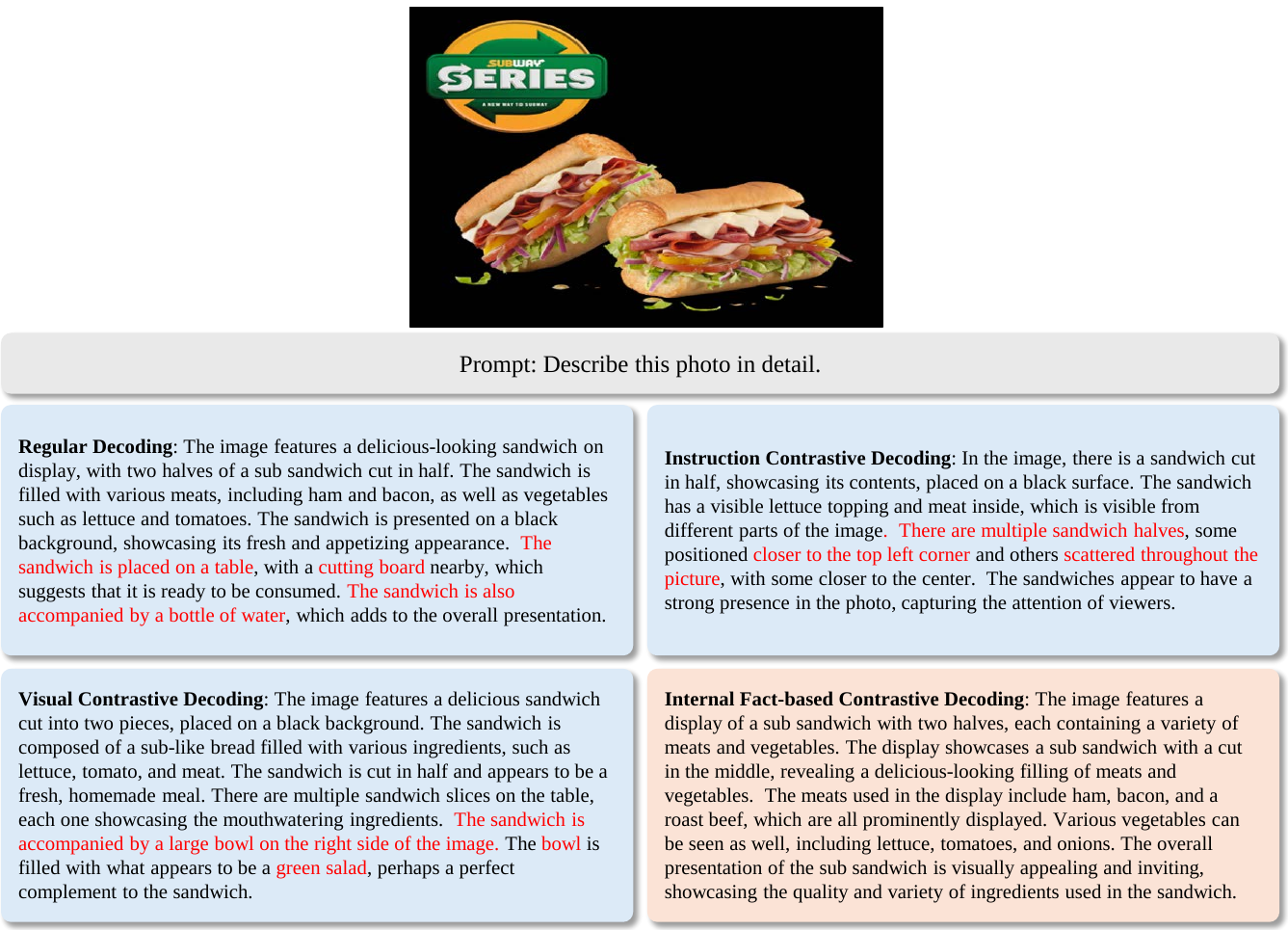}}
    \subfigure{
    \includegraphics[width=0.8\linewidth]{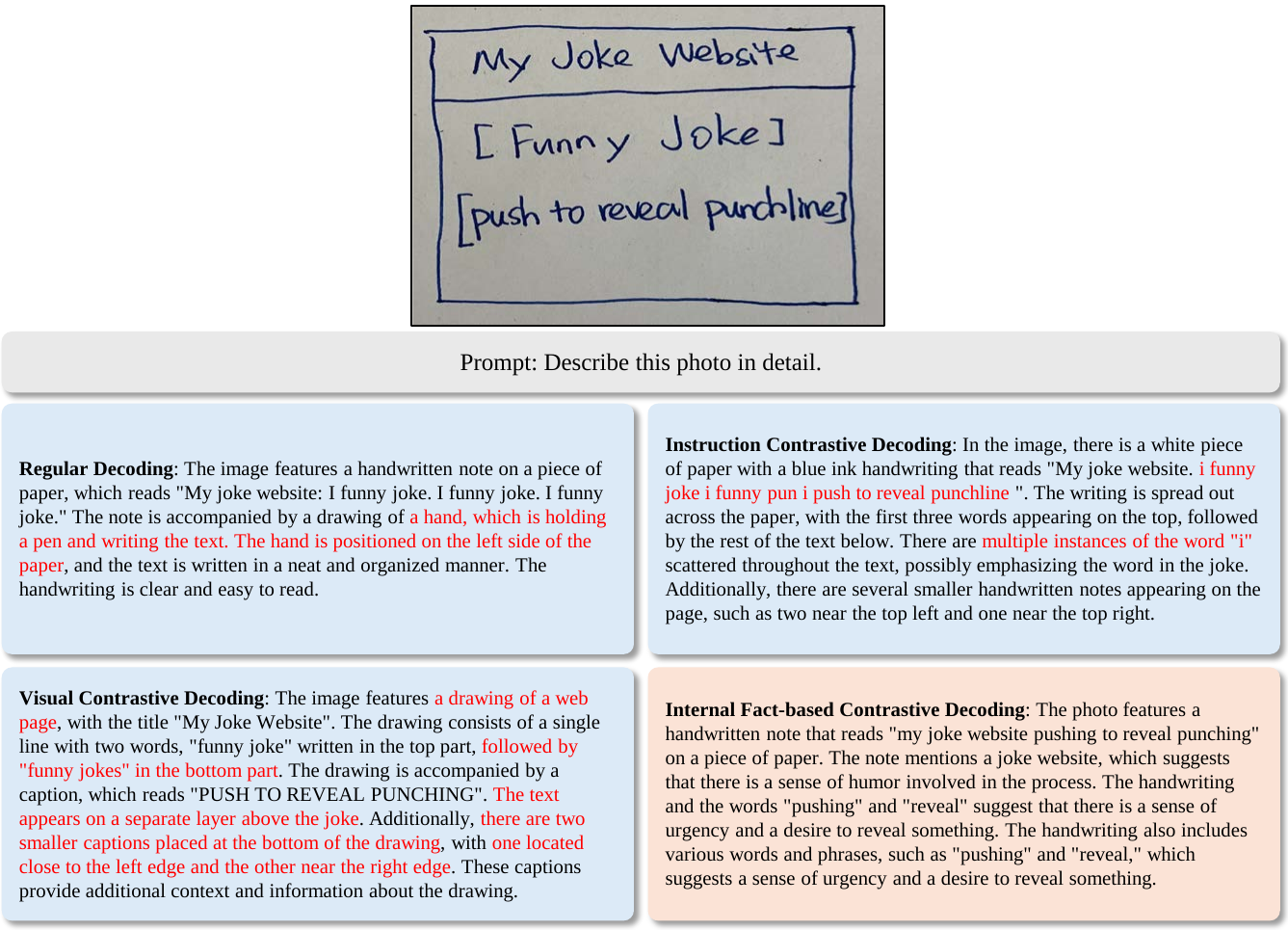}
    }
    \end{center}
    \vskip -0.2in
    \caption{LLaVA-Bench results comparing direct decoding, ICD, VCD, and IFCD with InstructBLIP backbone.}
    \label{fig: llava_bench_case_apex_blip}
\end{figure}
\begin{figure}[ht]
\vskip 0.2in
\begin{center}
    \subfigure{
    \includegraphics[width=0.8\linewidth]{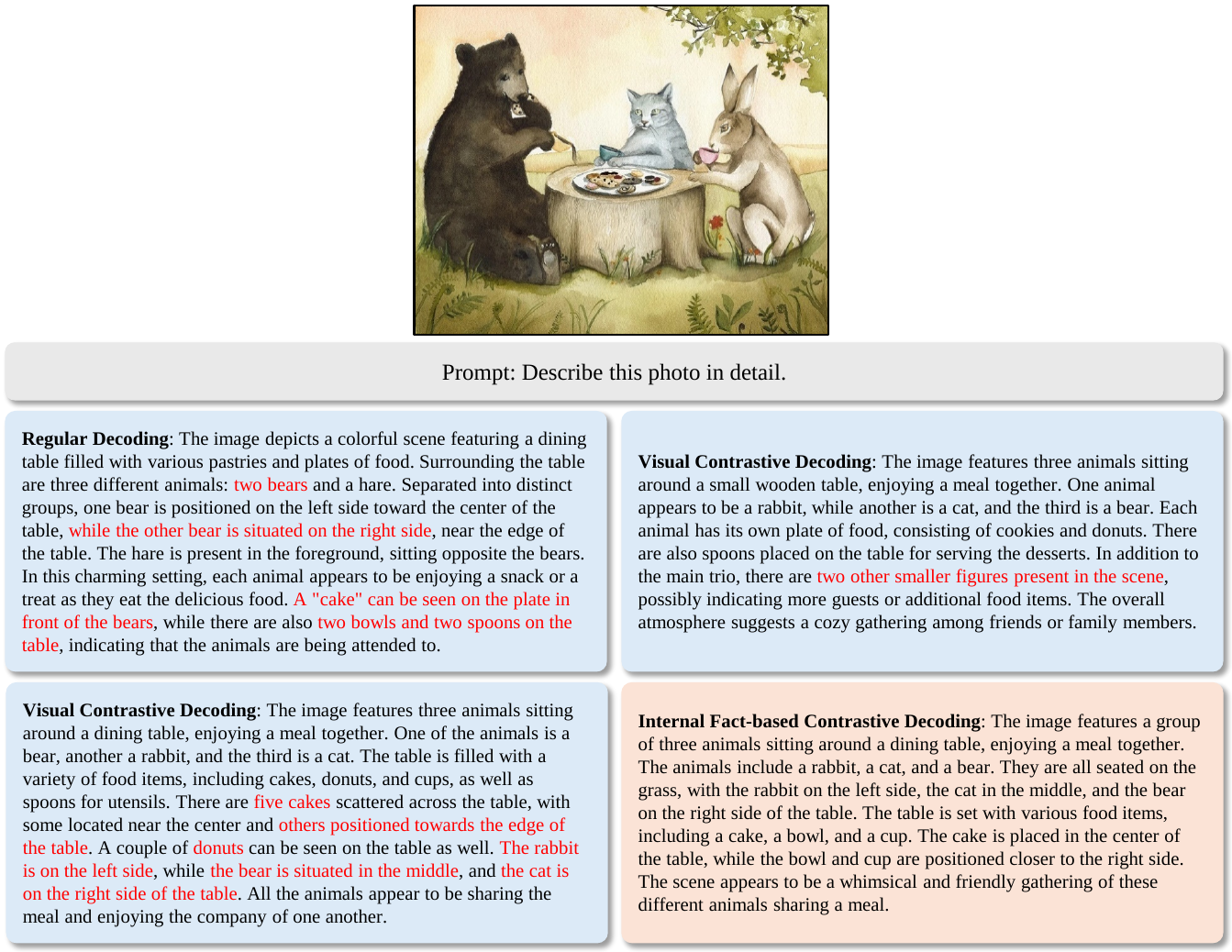}}
    \subfigure{
    \includegraphics[width=0.8\linewidth]{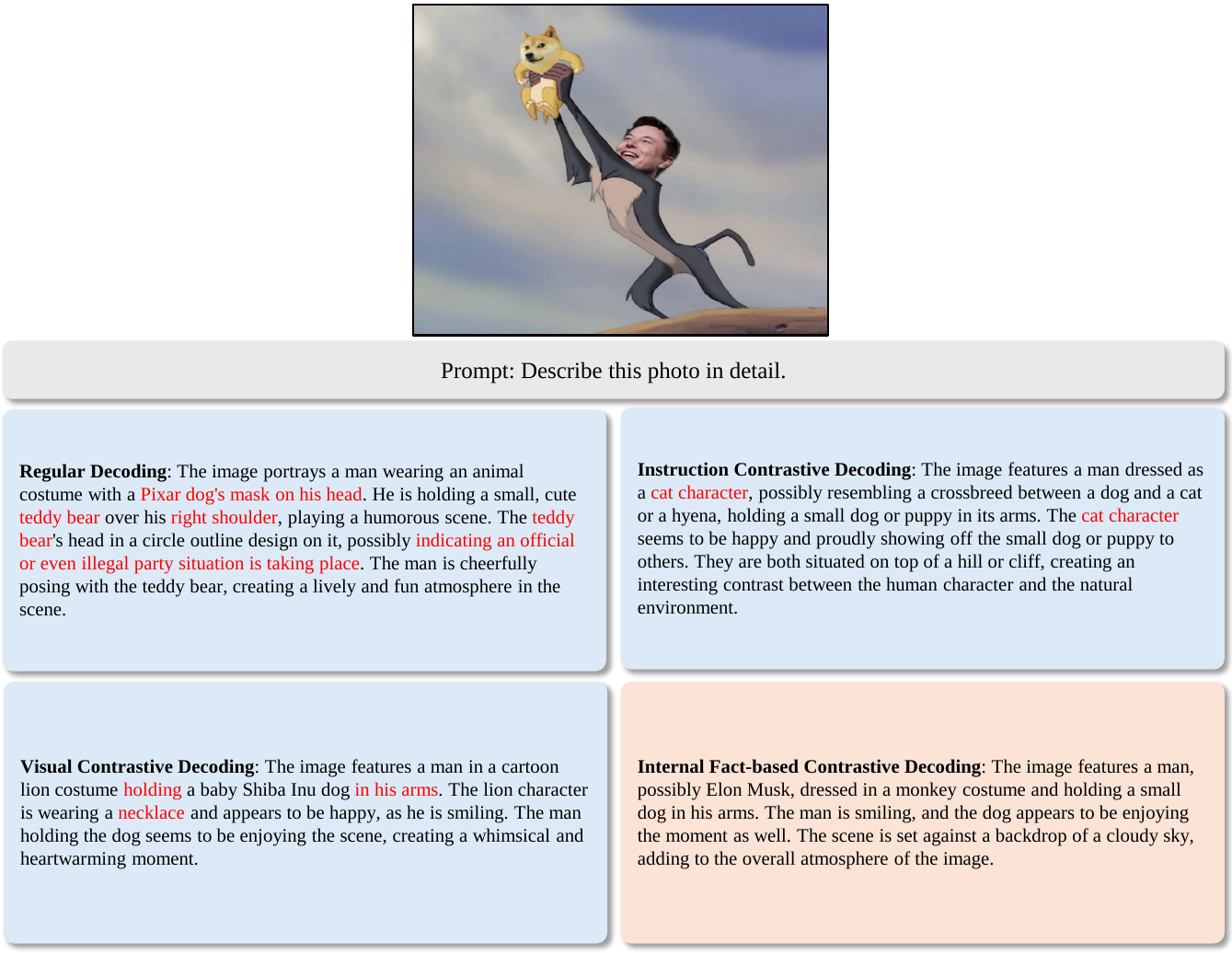}
    }
\end{center}
\vskip -0.2in
    \caption{LLaVA-Bench results comparing direct decoding, ICD, VCD, and IFCD with LLaVA 1.5 backbone.}
    \label{fig: llava_bench_case_apex_llava}
\end{figure}


\end{document}